\pdfoutput=1

\documentclass[journal, twocolumn]{IEEEtran}
\usepackage{amsmath,amsfonts}
\usepackage{algorithmic}

\usepackage{algorithm}
\usepackage{multirow}
\usepackage[numbers,sort]{natbib}
\usepackage{amsthm}
\usepackage{amssymb}
\usepackage{subfigure}
\usepackage[center]{caption}

\usepackage{lipsum}

\usepackage{array}
\usepackage[caption=false,font=normalsize,labelfont=sf,textfont=sf]{subfig}
\usepackage{textcomp}
\usepackage{stfloats}
\usepackage{float}
\usepackage{url}
\usepackage{verbatim}
\usepackage{graphicx}

\def\BibTeX{{\rm B\kern-.05em{\sc i\kern-.025em b}\kern-.08em
T\kern-.1667em\lower.7ex\hbox{E}\kern-.125emX}}
\usepackage{balance}

\def\b#1{\mathbf{#1}}
\def\t#1{\textbf{#1}}
\def\r#1{\mathrm{#1}}
\def\c#1{\mathcal{#1}}

\def\eqta{\begin{equation}}
\def\eqtb{\end{equation}}
\def\alna{\begin{aligned}}
\def\alnb{\end{aligned}}
\def\arra{\begin{array}}
\def\arrb{\end{array}}
\IEEEoverridecommandlockouts
\begin{document}

%header
\markboth{Journal of \LaTeX\ Class Files,~Vol.~18, No.~9, September~2020} %%
{How to Use the IEEEtran \LaTeX \ Templates}

\title{Clustering Based on Density Propagation and Subcluster Merging}
\author{Feiping Nie, \IEEEmembership{Senior Member, IEEE}, Yitao Song, Jingjing Xue, Rong Wang, and Xuelong Li, \IEEEmembership{Fellow, IEEE}

\thanks{The authors are are with the School of Artificial Intelligence, OPtics and ElectroNics (iOPEN), School of Computer Science, Northwestern Polytechnical University, Xi’an 710072, P.R. China, and also with the Key Laboratory of Intelligent Interaction and Applications (Northwestern Polytechnical University), Ministry of Industry and Information Technology, Xi’an 710072, P.R. China. (email: feipingnie@gmail.com; tombacsong@outlook.com; jingjing\_xue@mail.nwpu.edu.cn; wangrong07@tsinghua.org.cn; li@nwpu.edu.cn).}}

\markboth{Journal of \LaTeX\ Class Files,~Vol.~18, No.~9, September~2021}%
{How to Use the IEEEtran \LaTeX \ Templates}

%\twocolumn[
%#\begin{@twocolumnfalse}
\maketitle % important!
%\end{@twocolumnfalse}

%abstract
\begin{abstract}
We propose the DPSM method, a density-based node clustering approach that automatically determines the number of clusters and can be applied in both data space and graph space. Unlike traditional density-based clustering methods, which necessitate calculating the distance between any two nodes, our proposed technique determines density through a propagation process, thereby making it suitable for a graph space. 
In DPSM, nodes are partitioned into small clusters based on propagated density. The partitioning technique has been proved to be sound and complete. 
We then extend the concept of spectral clustering from individual nodes to these small clusters, while introducing the CluCut measure to guide cluster merging. This measure is modified in various ways to account for cluster properties, thus provides guidance on when to terminate the merging process. 
Various experiments have validated the effectiveness of DOSM and the accuracy of these conclusions.
\end{abstract}

%]

\maketitle

%Introduction
%1、聚类、图节点聚类的应用，研究意义
%2、现有方法及其劣势，k-means、谱聚类
%3、Motivation：高阶度？舍弃稀疏点和噪声点
%4、贡献：提出算法，权重传播的意义，实验
\section{Introduction}
\IEEEPARstart{I}{n} contemporary production and daily life, an increasing volume of data is generated, collected, inferred, and analyzed. Clustering, as a data mining technique, effectively analyzes the relationships inherent in the data and uncovers underlying knowledge and rules without supervision by categorizing similar samples into the same group while distinguishing dissimilar samples into separate categories\cite{pedrycz2005knowledge}. In numerous applications, including customer segmentation\cite{kansal2018customer,domingos2001mining}, recommendation systems\cite{li2012using,shao2009music}, anomaly detection\cite{moshtaghi2011clustering}, natural language processing\cite{probierz2022clustering}, and others—along with the upstream tasks associated with these applications—clustering methods have played a crucial role.

The current mainstream clustering algorithms can be broadly categorized based on their design principles into several types: partitioning methods, hierarchical methods, objective-based method, density-based methods, and others\cite{jain1999data}. Furthermore, depending on the nature of the data utilized, these algorithms can be classified as either extracting sample information from the data space or transforming sample information into a graph structure—this latter approach is commonly referred to as the community discovery task\cite{newman2004fast, wang2011community}.

Partitioning clustering methods, such as $k$-means\cite{lloyd1982least} and its enhancements\cite{nie2021coordinate, wang2022discrete}, initially identify several cluster centers and subsequently assign each data node into clusters based on specific criteria. These methods often struggle with clustering tasks involving irregularly shaped clusters, such as concentric circle datasets.

Hierarchical clustering methods, such as BIRCH\cite{zhang1996birch}, utilize a tree-like structure and can be categorized into two ideas. The first idea is divisive hierarchical clustering, which divides a cluster into two or more sub-clusters in a top-down manner, recursively continuing this process until the desired clustering objective is achieved, such as BKHK\cite{nie2020unsupervised}. The second idea is agglomerative hierarchical clustering, which merges two or more clusters from the bottom up through recursive processes\cite{szekely2005hierarchical}. A significant challenge for them lies in the irreversibility of recursive processing, which renders them sensitive to noise and outliers. Additionally, determining the optimal number of clusters remains a complex issue.

Objective-based approaches, including spectral clustering \cite{ng2001spectral, shi2000normalized}, categorize nodes by optimizing a predefined objective function. This reliance on an objective function typically necessitates knowledge of the number of classes in advance, posing challenges for tasks where the number of class clusters is uncertain or only an approximate range is known.

Density-based methods like density peaks clustering\cite{rodriguez2014clustering} and its variants\cite{xu2018dpcg, ren2020effective, lotfi2020density} determine clusters through node density calculations. However, current techniques for calculating node density generally require assessing relationships between all pairs of nodes, which complicates their application to graph-structured data. To address this limitation, we propose a novel density measure based on the propagation process that only requires evaluating relationships between each node and its local neighbors. Leveraging this new density metric allows DPSM to partition multiple small clusters that can then be merged following agglomerative hierarchical clustering principles. In the merging phase, we also draw inspiration from spectral clustering concepts and introduce the $CluCut$ measure to guide the merging process according to cluster structures.

The primary contributions of this paper are as follows:

1) Our proposed method, clustering based on Density Propagation and Subcluster Merging (DPSM), demonstrates remarkable clustering performance based on density while relying exclusively on local neighborhoods. It is capable of yielding favorable results in scenarios involving both an unknown number of clusters and a predetermined number.

2) The proposed propagation-based density detection technique effectively extracts density information from the relationships among nodes, demonstrating its efficacy in both data space and graph space in Section \ref{Density Propagation}. 

3)The technique of node partitioning, grounded in a strict partial order relation of density, ensures that the resulting small clusters are situated within actual clusters rather than spanning multiple real clusters. Additionally, we provide a completeness proof for this method in Section \ref{partition}.

4) Using spectral clustering as a reference, we propose the $CluCut$ measure for the task of cluster merging in Section \ref{Merge Clusters}. This measure is influenced by both intra-cluster and inter-cluster relationships, exhibiting distinct variations when compared to analogous properties in Ncut\cite{shi2000normalized}. Consequently, it effectively facilitates the merging of small clusters and enables DPSM to be applied without a predetermined number of classes.

5) Various experiments have corroborated the aforementioned four conclusions in Section \ref{sec:Experiment}.

The remainder of this article is organized as follows. In Section \ref{sec:Related}, we present several related methods that will serve as a basis for the detailed comparisons made later in this paper. Section \ref{sec:Proposed} provides an in-depth discussion of DPSM's flow and establishes its completeness through rigorous proof. In Section \ref{sec:Compare}, we analyze the differences and similarities between DPSM and related approaches, focusing on interpretation, implementation, and application scenarios. The experiments conducted in Section \ref{sec:Experiment} demonstrate the effectiveness of DPSM relative to similar algorithms, thereby validating the correctness of our conclusions. Finally, we conclude with a summary of findings and outline potential directions for future work in the final section.

%Related work
%1、传播过程？高阶的度？：刻画节点关系的过程
%2、DPC：密度峰值聚类
%3、谱聚类
\section{Related Work} \label{sec:Related}
\subsection{Propagation Procedures for Graph Learning} \label{Propagation}
\noindent In the field of graph learning, propagation serves as a widely adopted processing technique that transforms abstract and static graph structures into tangible and dynamic information flow processes. 

In traditional machine learning approaches, such as Label Propagation Algorithm (LPA)\cite{zhu2002learning, raghavan2007near} or Personalized PageRank (PPR)\cite{page1999pagerank}, the label information of nodes is propagated to achieve classification or significance of nodes. The algorithmic workflow can be uniformly articulated as follows:
\eqta
\left\{
\alna
& \b f^{(t+1)} = (1-\alpha)\b P \b f^{(t)} + \alpha \b f^{(0)}, t \geq 0, \\
& \b f^{(0)} \text{ is given,}
\alnb
\right.
\eqtb
where $\b f$ is a column vector of size $n$ that characterizes an attribute of these $n$ nodes, while $\b P$ is a propagation matrix of dimensions $n \times n$, which represents the mode of attribute propagation across the graph and $\alpha$ is a given parameter between 0 and 1. 

In LPA, the propagation matrix is defined as a normalized adjacent matrix, specifically expressed as $\b D^{-1/2} \b A \b D^{-1/2}$ where $\b A$ denotes the adjacent matrix and $\b D$ is a diagonal matrix composed of the degrees. The attribute vector $\b f$ encodes the label information, with its initial value being a vector that assigns $1$ to labeled components and $0$ to unlabeled components. In PPR, the matrix $\b P$ characterizes the relationships between pairs of pages, while $\b f^{(0)}$ represents the interests of specific users or particular topics.

A comparable framework is employed for feature propagation in Graph Convolutional Networks(GCN)\cite{defferrard2016convolutional}:
\eqta
\left\{
\alna
& \b H^{(t+1)} = \sigma(\tilde{\b P} \b H^{(t)} \b W), t \geq 0, \\
& \b H^{(0)} = \b X,
\alnb
\right.
\eqtb
where $\b X$ denotes the input features, $\b H$ signifies the features obtained after integrating neighboring information, and the propagation matrix $\tilde{\b P}$ is typically defines as $(\b I + \b D)^{-1/2}(\b I + \b A)(\b I + \b D)^{-1/2}$. Here, $\b W$ represents the weights to be trained, and $\sigma(\cdot)$ refers to the activation function.

We have gained insights from the propagation procedures utilized in these methods; however, we did not propagate label or feature information. Our focus is on the extraction of graph structure during the propagation procedures, which allows us to derive node density information accordingly, as elaborated in Section \ref{Density Propagation}.

\subsection{Density Peak Clustering} \label{DPC}
\noindent Density Peak Clustering(DPC)\cite{rodriguez2014clustering}  is a density-based clustering method that identifies clusters by locating regions of higher density. In contrast to traditional K-means or hierarchical clustering approaches, DPC aims to discover naturally occurring high-density regional centers that act as cluster centroids. This methodology is particularly well-suited for addressing non-convex data distributions.

The prevailing  definition of node density in DPC is based on the Euclidean distance between any two nodes, as shown below:
\eqta
%\rho_i = \sum_j \exp(- \frac{d_{ij}^2}{d_c^2}),
\rho_i = \sum_j \chi(d_{ij} - d_{c}),
\eqtb
where $\chi(x) = 1$ if $x<0$ and $\chi(x) = 0$ otherwise, $d_{ij}$ denotes the Euclidean distance between node $i$ and node $j$, and the cutoff distance $d_c$ is a parameter that needs to be adjusted.

Based on the densities of nodes, DPC is capable of calculating the relative distance $\delta$ for each node, as illustrated below:
\eqta
\delta_i = 
\left\{
\alna
\min_j (d_{ij}), & \ \ \ j: \rho_j > \rho_i, \\
\max_j (d_{ij}), & \ \ \ \text{otherwise.}
\alnb
\right.
\eqtb

For any cluster, the node exhibiting the highest density is typically identified as the cluster center. Should there exists another node with a higher density, it must belong to a different cluster, thereby ensuring that the relative distance of the original cluster center remains large. Consequently, for each cluster, there exists one and only one node with a significant relative distance. Leveraging this property, DPC can identify nodes with both high density and considerable relative distance as clustering centers, subsequently facilitating the determination of clusters.

DPC provides significant inspiration for considering clustering from a density perspective. However, it necessitates the computation of distances between a large number of nodes, leading to unacceptable time complexity in large datasets. The Density Peak Clustering Algorithm Based on Grid (DPCG) \cite{xu2018dpcg} employs grid structures to alleviate this issue. In contrast, we prefer to restrict distance calculations to the local neighborhood, which not only reduces computational complexity but also fits more manifold structures.

We place greater emphasis on the local neighborhood,which means that cluster centers cannot be determined through direct global density comparisons. In a manner similar to the density peaks clustering algorithm based on layered k-nearest neighbors and subcluster merging (LKSM-DPC) \cite{ren2020effective}, we incorporate merging procedures into our approach. However, while LKSM-DPC draws inspiration from the law of gravitation and typically merges two adjacent large clusters, DPSM is informed by spectral clustering principles and tends to merge two adjacent small clusters.

\subsection{Spectral Clustering} \label{Spectral Clustering}
\noindent Spectral Clustering (SC)\cite{ng2001spectral} is a classic graph theory-based clustering method that seeks to uncover the intrinsic structure of data by examining the feature vectors derived from the Laplacian matrix. One of its physical interpretations is to identify the optimal cut of a graph. Taking the NCut method\cite{shi2000normalized} as an example, Ncut between connected block $A$ and connected block $B$ can be expressed as follows:
\eqta
NCut(A,B) = \frac{cut(A,B)}{vol(A)} + \frac{cut(A,B)}{vol(B)},
\eqtb
where $cut(A, B)$ denotes the total weight of edges connecting $A$ and $B$, while $vol(A)$ represents the total degree of nodes within $A$:
\eqta \label{def:cut}
cut(A, B) := \sum_{(u,v)\in \c E: u \in A, v \in B} w(u,v),
\eqtb
\eqta \label{def:vol}
vol(A) := \sum_{v \in A} deg(v) = \sum_{v \in A, (u,v)\in \c E} w(u,v).
\eqtb

This design ensures that the connections between the blocks are minimized, while maintaining proximity among the nodes within each connected block. Similar criteria is applied to cluster merging by  defining inter-cluster relationships as well as intra-cluster relationships, as illustrated in Section \ref{Merge Clusters}.

\begin{figure*}[t!]
\centering
\begin{tabular}{@{}*{2}{p{9cm}}@{}} % 6列居中对齐
    \includegraphics[width=0.5\textwidth]{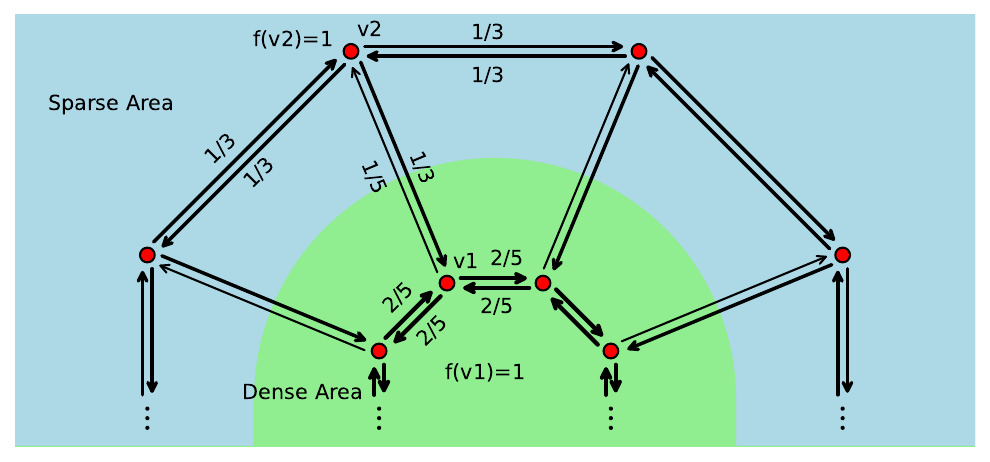} &
    \includegraphics[width=0.5\textwidth]{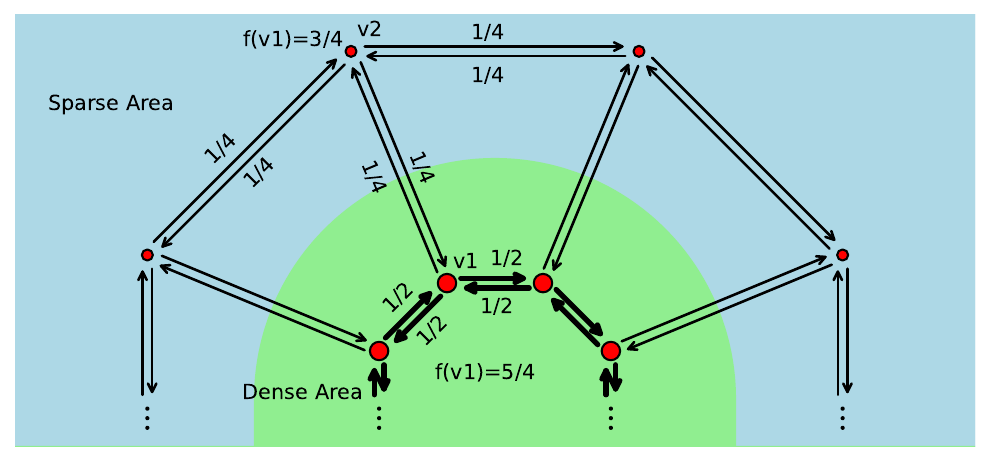} \\
(a) \t{Initail:} $v_1$ in the dense area and $v_2$ in the sparse area initially possess identical degrees of $1$; however, the degrees propagated between them are not equal, that is, $1/5 < 1/3$. 
& (b) \t{Steady States:} Over time, the degree of $v_1$ approaches $5/4$ while that of $v_2$ approaches $3/4$. This ensures consistency in the degrees  propagated between them, that is, $1/4 = 1/4$.
\end{tabular}
\caption{Schematic diagram of density propagation}
\label{Fig:Propagation}
\end{figure*}

%Proposed Method
%1、权重传播过程->表述密度
%2、根据权重传播确定节点类型
%3、合并策略
\section{Proposed Method, DPSM} \label{sec:Proposed}

\newtheorem{lemma}{Lemma}
\newtheorem{theorem}{Theorem}
\newtheorem{definition}{Notation}
\newtheorem{corollary}{Corollary}
\newtheorem{property}{Property}
\renewcommand{\qedsymbol}{$\blacksquare$}

\noindent In this section, we present the details of Clustering based on Density Propagation and Subcluster Merging (DPSM), which is organized into three main components. Firstly, we utilize the weighted adjacent matrix of graph nodes as a propagation paradigm to derive the densities of each node. This approach requires fewer hyperparameters and can be effectively adapted to manifolds of varying scales. Secondly, the propagated densities manifest as multiple peaks within the space, thereby facilitating natural clustering. Finally, when the number of clusters exceeds expectations or when distinctions between certain clusters are not clear-cut, several clusters are merged and updated sequentially.

%1、权重传播过程->表述密度；好处是？
% 近邻关系·稀疏矩阵，流形尺度无关
\subsection{Evaluating Nodes by Density Propagation} \label{Density Propagation}
\noindent Density is a crucial property in the topology of graphs. However, when calculating node densities, it often becomes necessary to specify certain parameters and challenging to achieve consistent density measurements across graphs of varying scales. To address this issue, we propose a propagation procedure for obtaining densities.

Consider a graph $\c G = (\c V, \c E, w)$ where $\c V$ represents the set of $n$ nodes, $\c E \subset \c V \times \c V$ denotes the set of edges and $w: \c E \rightarrow \mathbb R^{+}$ indicates the mapping from edges to their respective weights.

We aim for the dense degrees to propagate across the graph over an extended period; therefore, it is essential to normalize the outgoing edge weights of each node. Denote $p_{ij}$ as the normalized edge weight from node $j$ to node $i$, which also represents the propagation weight:
\eqta
p_{ij} = \left\{
\alna
\frac{w(j,i)}{\displaystyle \sum_{(v_j, v_t) \in \c E}w(j,t)}, &&(v_j, v_i) \in \c E, \\
0, &&\r{otherwise},
\alnb
\right.
\eqtb
thus the normalized weight matrix $\b P \in [0,1] ^{n \times n}$, constructed from $p_{ij}$ is typically sparse. Each column of $\b P$ sums to $1$ and all diagonal elements of $\b P$ is all $0$.

At the outset, each node possesses a dense degree $f$ of $1$. These degrees propagate synchronously throughout the graph in accordance with $p$. If we concatenate the degrees $f$ of all nodes into a row vector $\b f \in (\mathbb R^{+})^{n \times 1} $, we have:
\eqta \label{prop}
\left\{
\alna
& \b f^{(t+1)} = \b P \b f^{(t)}, t \geq 0, \\
& \b f^{(0)} = \b 1,
\alnb
\right.
\eqtb
where $\b f$ denotes the degrees of nodes, $t$ denotes the number of iterations and $\b 1$ denotes the column vector of ones. 

In each iteration, the degree propagation between nodes in both dense and sparse areas remains relatively stable. However, when a node in a sparse area connects to a node in a dense area, the latter shows a greater tendency to propagate its degree to neighboring nodes within the dense area rather than those in the sparse area. As illustrated in Fig. \ref{Fig:Propagation}, this phenomenon results in a spontaneous shift of degrees from the sparse area to the dense area, thereby serving as an indicator of density. 

\begin{figure}[t]
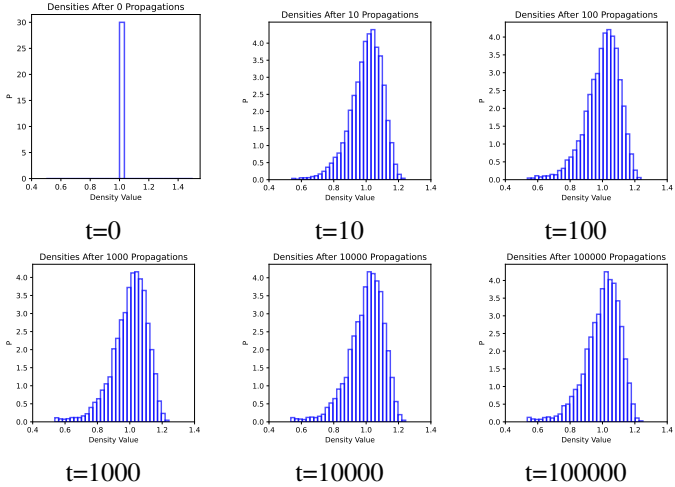

\centering
\begin{tabular}{@{}*{6}{c}@{}} % 6列居中对齐
    \includegraphics[width=0.15\textwidth]{figures/Ours\_iter0.pdf} &
    \includegraphics[width=0.15\textwidth]{figures/Ours\_iter10.pdf} &
    \includegraphics[width=0.15\textwidth]{figures/Ours\_iter100.pdf} \\
    t=0 & t=10 & t=100 \\
    \includegraphics[width=0.15\textwidth]{figures/Ours\_iter1000.pdf} &
    \includegraphics[width=0.15\textwidth]{figures/Ours\_iter10000.pdf} &
    \includegraphics[width=0.15\textwidth]{figures/Ours\_iter100000.pdf} \\
    t=1000 & t=10000 & t=100000
\end{tabular}
\caption{The Probability Density Distribution of Density After Specific Iterations}
\label{Fig:Dist}
\end{figure}

After multiple iterations, the steady-state density degrees manifest as a distribution. As illustrated in Fig. \ref{Fig:Dist}, using the USPS dataset\cite{Hull1994ADF} as an example, the probability density distribution of node density remains stable after hundreds of iterations. Furthermore, we focus on relative relationships between densities rather than their absolute values; thus, we can utilize the density obtained after hundreds of iterations as a reliable guide to achieve consistent clustering results while minimizing computational overhead.

A node with a lower dense degree is typically situated at the outskirt of a cluster or serves as the boundary between two or more clusters. Our objective is to identify and eliminate nodes located along these dividing lines, which necessitates considering the graph in its entirety.

%2、根据权重传播确定节点类型
% 低密度意味着簇的分界，高密度意味着簇的中心。从高密度到低密度自然成簇。
% 节点类型
\subsection{Partitioning Nodes With Propagated Densities} \label{partition}
\noindent Instead of using the absolute values of density degrees, we utilize them in accordance with the strict partial order relationship of the neighborhood of each node. For example, in the context of density space, a local peak node whose neighboring nodes all exhibit lower densities is necessarily considered the root of a cluster, which subsequently expands around it. During this expansion, if all neighbors with higher densities of a node belong to the cluster, there is a high probability that this particular node also belongs to the cluster. 

The strict partial order relation concerning density plays a pivotal role in the partitioning of nodes within this section. Consequently, we redefine the propagated density $\rho$ based on the density $\b f$, ensuring that the propagated densities for each pair of nodes are distinct:
\eqta
\forall v_i, v_j \in \c V:
\left\{
\alna
\rho(v_i) > \rho(v_j), \text{if } (f_i > f_j) \lor (f_i = f_j \land i < j), \\
\rho(v_i) < \rho(v_j), \text{if } (f_i < f_j) \lor (f_i = f_j \land i > j),
\alnb
\right.
\eqtb
where $f_i$ represents the $i$-th element of the dense degree vector $\b f^{(t)}$ after $t$ iterations, while $\rho(v_i)$ signifies the propagated density of the node $v_i$.

\begin{definition}[] Given that the densities are distinct, all peaks are grouped into the root set $\c V_\r{Rt}$ and referred to as the root nodes.
\eqta
\c V_\r{Rt} := \{v_i \in \c V \mid \forall (v_i,v_j) \in \c E: \rho(v_i) > \rho(v_j)\}
\eqtb
\end{definition}

\begin{definition}[] Each root node $r$ in $\c V_\r{Rt}$ corresponds to a cluster set denoted as $\c C(r)$, which not only includes the root, but also encompasses other nodes that can be reached through monotone descent from the root $r$. The cluster $\c C(r)$ represents the limit of a recurrence which converges due to the finite number of nodes.
\eqta
\alna
& \c C(r)^{(0)} = \{r\},\ r\in \c V_\r{Rt}, \\
& \c C(r)^{(t+1)} = \c C(r)^{(t)} \cup \{v_i \in \c V \mid\\
&\ \ \ (\forall (v_i,v_j) \in \c E \text{ s.t. } \rho(v_j) > \rho(v_i) \implies v_j \in \c C(r)^{(t)})\\
&\ \ \ \land(\exists (v_i,v_j) \in \c E \text{ s.t. } \rho(v_j) > \rho(v_i))\}, t>0, \\
& \c C(r) := \lim_{t \rightarrow +\infty}\c C(r)^{(t)},\ r\in \c V_\r{Rt}.
\alnb
\eqtb
\end{definition}

\begin{definition}[] To separate two or more clusters, we define the margin set $\c M(r)$ for each cluster $\c C(r)$ associated with the root node $r$. This approach is employed to address boundary cases, and we refer to the nodes situated within any margin sets as margin nodes.
\eqta
\c M(r) := \{v_i \in \c V \mid \exists (v_i,v_j) \in \c E \text{ s.t. } v_j \in \c C(r)\} \setminus \c C(r).
\eqtb
\end{definition}

% \vspace{1em}
With these notations, we can derive the following conclusions.

\begin{property}[]
A node within any cluster cannot simultaneously belong to another cluster.
\end{property}
\begin{proof}
1) Let us suppose a cluster $\c C(r)$ whose root is $r$, containing $t$ nodes. By ordering the nodes within the cluster according to their density $\rho$, from highest to lowest, we obtain a sequence: $(r, c_2, c_3, ..., c_t)$, where $\rho(r) > \rho(c_2) > \rho(c_3) > \ldots > \rho(c_t)$.

Firstly, it follows that $r$ does not belong to any other clusters. Secondly, for any positive integer $k < t$, if the first $k$ nodes in this sequence do not belong to any other clusters, then since all higher neighbors of $c_{k+1}$ are among them, this node will also not be included in any other clusters.

By employing mathematical induction, we establish the validity of this conclusion.
\end{proof}

\begin{property}[]
A node in any margin set cannot be a member of a cluster, and vice versa.
\end{property}
\begin{proof}
Assuming $m$ is a node in $\c M(r)$, by definition, there exists at least one node $c$ that belongs to $\c C(r)$ and is adjacent to $m$. If $\rho(m)$ is greater than $\rho(c)$, then it follows paradoxically that $c$ should not belong to $\c C(r)$. Therefore, when the densities are distinct, it must be the case that $\rho(m) < \rho(c)$.
 
A high neighbor $c$, which belongs to $\c C(r)$ renders it impossible for node $m$ to belong to any other clusters. Since $m$ also does not belong to $\c C(r)$, this implies that $m$ cannot be a member of any clusters.

As demonstrated in \t{Property 1}, all higher neighbors of a node $c$ in $\c C(r)$ are not part of any other cluster. Simultaneously, all lower neighbors also do not belong to any other cluster, indicating that it cannot serve as the margin node for any other cluster. Consequently, it is excluded from belonging to any margin set.
\end{proof}

\begin{figure*}[t]
\centering
\begin{tabular}{@{}*{2}{p{9cm}}@{}} % 2列居中对齐
\includegraphics[width=0.5\textwidth]{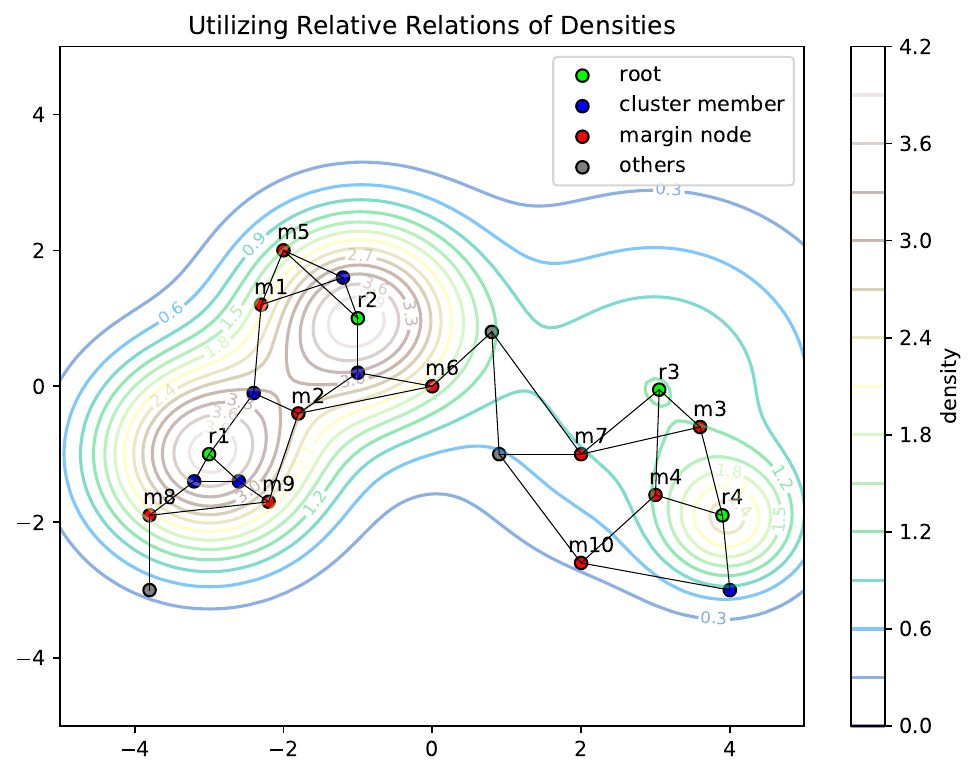} &
\includegraphics[width=0.5\textwidth]{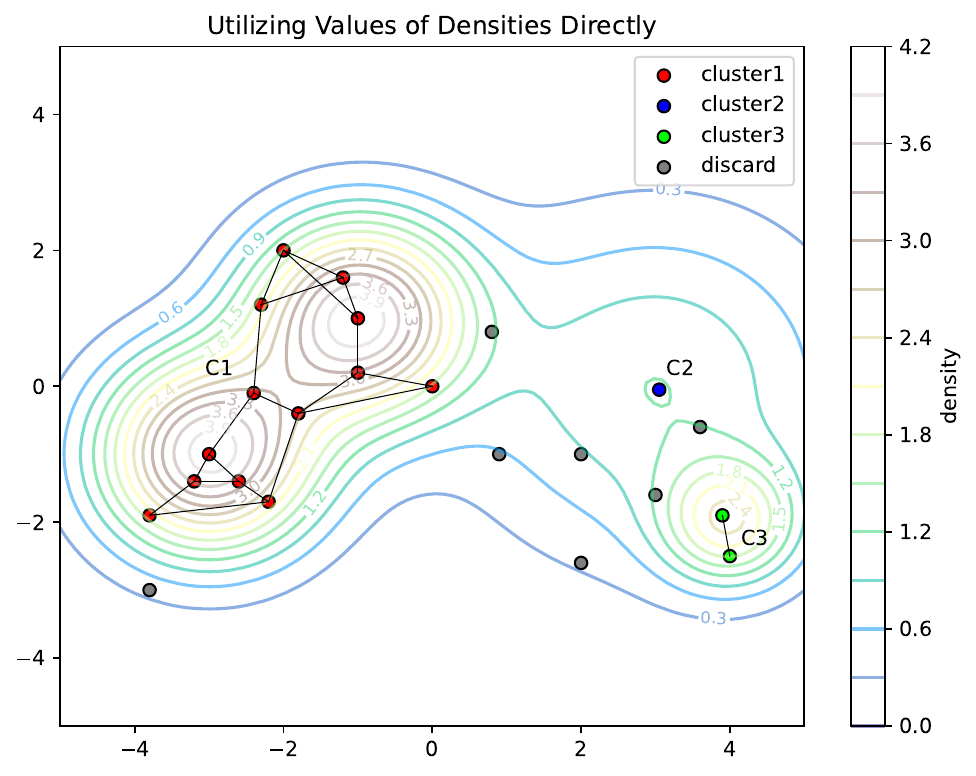} \\
(a) The contours illustrate trends in density. The green nodes (from $r_1$ to $r_4$) are local peaks and thus classified as root nodes. The blue nodes exhibit monotonically descending patterns from the green nodes, with all higher neighbors belonging to the same cluster; therefore, they can be identified as part of this cluster. Nodes $m_1$, $m_2$, $m_3$, and $m_4$ simultaneously connect two clusters, categorizing them as margin nodes, while nodes $m_5$ to $m_9$ possess higher margin neighbors that complicate the determination of their ownership; nonetheless, they are also designated as margin nodes. The remaining nodes are depicted in gray. \t{Our method, DPSM, applies this approach.}&  
(b) Using the density value directly implies that the classification of nodes is based on a specific threshold. When the density of a node surpasses the threshold, it is retained; conversely, if it does not meet this criterion, it is discarded. When identifying three representative clusters, a density boundary of approximately $1.3$ is established. After discarding all nodes with densities lower than $1.3$, the remaining nodes form three connected blocks: $\c C_1$, $\c C_2$, and $\c C_3$. However, one might consider that cluster $\c C_1$ should be further divided into two distinct clusters due to the gap present in its middle. In contrast, clusters $\c C_2$ and $\c C_3$ may be regarded as a single entity.
\end{tabular}
\caption{Schematic Diagrams of the Density-based Node Partitioning Approaches.}
\label{Fig:Example}
\end{figure*}

\begin{property}[]
The path from a node in one cluster to a node in another must traverse at least one margin node.
\end{property}
\begin{proof}
Assume that the path from a node $a_0$ in $\c C(a)$ to a node $b_0$ in $\c C(b)$ can be represented as a sequence given by $(a_0, a_1, \ldots , a_{t_a}; u_1, u_2, \ldots, u_k; b_{t_b}, b_{t_b-1}, \ldots, b_1, b_0)$, where $a_0, a_1, \ldots, a_{t_a} \in \c C(a)$, $b_0, b_1, \ldots, b_{t_b} \in \c C(b)$ and $u_1,u_k \notin \c C(a) \cup \c C(b)$. This is ensured by \t{Property 1} and \t{Property 2}

Suppose there are no nodes $u_1, u_2, \ldots, u_k$, which implies that there is a direct edge connecting $a_{t_a}$ and $b_{t_b}$. Let’s assume $\rho(a_{t_a}) > \rho(b_{t_b})$. Under this condition, it follows that $b_{t_b}$ should not be included in the set $\c C(b)$. This leads to a contradiction. Therefore, we conclude that at least one node, denoted as $u_1$, must exists.

According to \t{Notation 3}, $u_1$ belongs to the margin set $\c M(a)$ and qualifies as a margin node.
\end{proof}

\begin{property}[]
There exists at least one margin node throughout the entire graph that belongs to two or more margin sets concurrently. 
\end{property}
\begin{proof}
Among all nodes that do not belong to any cluster, the node with the highest density is designated as $m$. Since $m$ is not a root node, it must have higher-density neighbors, which are members of some clusters.

If these neighbors all belong to the same cluster, then paradoxically, $m$ would also be within it. Therefore, it follows that these neighbors must belong to multiple clusters, indicating that $m$ belongs to multiple margin sets.
\end{proof}
\vspace{1em}

With these definitions in place, we can efficiently derive these sets by examining nodes in descending order of density. An illustrative example is provided in Fig. \ref{Fig:Example} (a). 

Additionally, there is a noteworthy detail to discuss. Utilizing the values of density degrees does not achieve the intended outcome, as demonstrated in Fig. \ref{Fig:Example} (b). If we directly utilize the values of density degrees and discard nodes with low density, we risk discarding a significant amount of valuable information when identifying clusters based on connection blocks. In contrast, by utilizing the relative relations of densities, we can retain this crucial information for subsequent merging processes.

\subsection{Merging Clusters via Inter- and Intra-cluster Relationships} \label{Merge Clusters}
\noindent After partitioning clusters and margin sets from $\c V$, it is highly probable that the nodes within the same cluster will belong to the same class. Typically, these clusters exceed the desired quantity or some of them may actually represent the same class, thus necessitating additional merging processes.

Based on \t{Property 3} and \t{Property 4}, it is evident that there exist nodes within the margin sets of multiple clusters simultaneously, indicating a higher likelihood that these clusters originate from the same class. We refer to such clusters as adjacent, provided that there are overlapping nodes in their respective margin sets:
\eqta
\c C(a) \text{ is adjacent to } \c C(b) \iff \c M(a) \cap \c M(b) \neq \varnothing.
\eqtb

Among these adjacent clusters, pairs characterized by a greater inter-cluster relationship and a lesser intra-cluster relationship are preferentially merged\cite{von2007tutorial}. Our definition of inter-cluster relationships pertains to the margin nodes, which is why we define margin sets. The inter-cluster relationships between the adjacent clusters $\c C(a)$ and $\c C(b)$ can be expressed as follows:
\eqta \label{R_inter}
\alna
&R_\r{inter}(\c C(a), \c C(b))  \\
:= &\sum_{m \in \c M(a) \cap \c M(b)} \min\{\sum_{c_a \in \c C(a)} w(m, c_a), \sum_{c_b \in \c C(b)} w(m, c_b)\}.
\alnb
\eqtb

Junction nodes $v$ connect two distinct clusters, with the relationship between these clusters represented by the cumulative contributions of such nodes. For each junction node $v$, there exist one or more edges associated with both clusters, respectively. Considering that it is probable to classify the node into the cluster exhibiting a stronger association, we propose utilizing its association with the cluster exhibiting a weaker connection as its contribution to the inter-cluster relationship.

The intra-cluster relationships encapsulate the comprehensive interactions among nodes within a cluster, thereby determining the significance of inter-cluster relationships to that cluster. The intra-cluster relationship of the cluster $\c C(r)$ is defined as follows:
\eqta \label{R_intra}
R_\r{intra}(\c C(r)) := \sum_{u, v \in \c C(r), (u, v)\in \c E} w(u, v).
\eqtb

In definitions (\ref{R_inter}) and (\ref{R_intra}), edge weights $w$ are employed instead of propagation weights $p$. This decision is based on the requirement for symmetric relationships between nodes in both inter- and intra-relationships, which is not fulfilled by propagation weights $p$.

Referring to the idea of SC, our objective is to minimize the sum of 'cuts' between clusters. We define $CluCut(a, b)$ as a measure of inter-cluster relationships in relation to intra-cluster relationships, which can be expressed as follows:
\eqta\label{CluCut}
CluCut(a,b) :=
\frac{R_\r{inter}(\c C(a), \c C(b))}{R_\r{intra}(\c C(a))} + \frac{R_\r{inter}(\c C(a), \c C(b))}{R_\r{intra}(\c C(b))}.
\eqtb

The objective can be written as:
\eqta
\min_{\c C_1, \c C_2, ..., \c C_k} \sum_{\substack{
a,b \in \c V_\r{Rt}, \\
\c M(a) \cap \c M(b) \neq \varnothing
}} CluCut(a,b).
\eqtb

To achieve this, the clusters $\c C(a)$ and $\c C(b)$ will be merged if the current total number of clusters exceeds the required amount and they meet the following conditions:
\eqta \label{merge}
(a, b) = \arg \max_{\substack{
a,b \in \c V_\r{Rt}, \\
\c M(a) \cap \c M(b) \neq \varnothing
}} CluCut(a, b).
\eqtb

The relatively minor intra-cluster relationships assume a more significant role, which can be interpreted as facilitating the integration of smaller clusters into larger ones.

Merging $\c C(a)$ and $\c C(b)$ involves straightforward processes:

1) Update $\c C(a)$ to present the union of both sets.

2) Remove node $b$ from the root set $\c V_\r{Rt}$, and eliminate the corresponding sets $\c C(b)$ and $\c M(b)$.

3) Adjust $\c M(a)$ in accordance with \t{Notation 3}. 

It can be readily demonstrated that the resulting graph continues to satisfy the established properties and are ready for further merging.

After several merging processes, we obtain multiple clusters that meet the specified requirement, along with a number of nodes that do not belong to any cluster. The membership status of these nodes remains ambiguous; therefore, one may either assign them to the nearest cluster or simply drop them off.

If the number of classes is not given, DPSM should terminate based on the change in the value of $CluCut$ that occurs with each merging event. When primary clusters have been merged, there will be a significant drop in the maximum $CluCut$.

When addressing outliers and small clusters, the intra-cluster relationships can serve as a valuable reference. During the merging process, we will eliminate certain clusters when their quantity and intra-cluster relationships are both significantly below average; for instance, when they fall below 5\% of the average.

%Discussion
%1、与密度峰值聚类比较
%2、与谱聚类比较
\section{Similarities and Differences With Other Algorithms} \label{sec:Compare}
\begin{figure}[H]
\centering
\caption{Comparison Between Implementations of Methods} \label{Fig:Compare}
\includegraphics[width=0.5\textwidth]{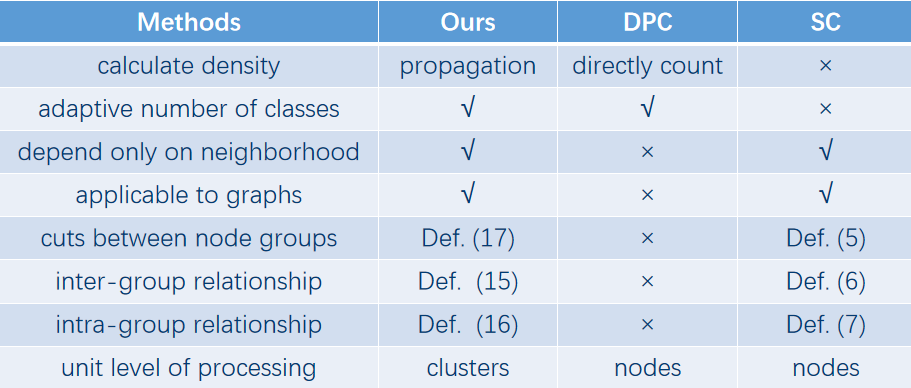}
\end{figure}

\noindent DPSM is inspired by DPC\cite{rodriguez2014clustering} and SC\cite{ng2001spectral}; however, it diverges significantly from their implementations. Refer to Fig. \ref{Fig:Compare} for a comparison of the main similarities and differences.

\subsection{Compared With DPC}
\noindent DPSM is also density-based, akin to DPC\cite{rodriguez2014clustering}, which enables it to effectively manage irregularly shaped clusters. 

For a data set where the number of clusters is unknown, DPC can ascertain both the number of cluster centers and the total number of clusters based on a specified threshold related to density $\rho$ and relative distance $\delta$. In contrast, DPSM determines the termination condition for cluster merging by monitoring changes in $CluCut$, as defined in Eq. (\ref{CluCut}). This allows us to select the most suitable option within a predetermined range for the number of clusters. As density-based methods, both of them are adept at producing clustering results without necessitating prior knowledge regarding the number of clusters.

DPSM computes density in a manner distinct from that of DPC. While DPC calculates density based on the distance between any two nodes, DPSM relies solely on the relationships among nearby nodes. Therfore, DPC encounters challenges with graph data, while DPSM effectively addresses these issues. 

When addressing a node set in space, DPC exhibits a time complexity of $O(n^{2}d)$, where $n$ represents the number of nodes and $d$ denotes the dimension. In contrast, the process of building a graph required by DPSM achieves a time complexity of $O(ndk \log n)$ when employing the $k$-nearest neighbor algorithm based on ball tree structures, with $k$ indicating the number of neighbors. The total time complexity of DPSM is $O(ndk \log n + nkt)$ which can be decomposed into 3 parts:

1) The complexity required for dimensionality reduction of data using Barnes-Hut t-SNE\cite{van2014accelerating} is $O(nd \log n)$. Then the construction of the graph results in a time complexity of $O(ndk \log n)$ due to the above discussion.

2) Obtaining density through propagation involves a time complexity of $O(nkt)$ where $t$ represents the number of propagation iterations, typically set at 100.This complexity arises because the propagation matrix $\b P$ in Eq. (\ref{prop}) is a sparse matrix with no more than $2nk$ elements.

3) In Section \ref{partition}, we clarify the properties of the node partitioning technique to ensure that each node is counted only once throughout the entire merging process. The time complexity associated with this technique approaches $O(n)$ by utilizing a disjoint set union\cite{galil1991data}. By leveraging margin nodes, we can employ a priority queue\cite{van1976design} to identify the two clusters that require merging, which results in a time complexity not exceeding $O(n \log n)$.

\subsection{Compared With Spectral Clustering}
\noindent In the third part of DPSM, the cluster merging process, we draw upon an interpretation of SC\cite{ng2001spectral}. SC operates by dividing a connected block into two segments when there are minimal connections between them and when the cohesion within each segment is high. In contrast, our method merges two small clusters into a single cluster when the inter-cluster relationship between these clusters is substantial while their respective intra-cluster relationships are relatively weak. 

The intra-cluster relationship $R_\r{intra}$ in our method bears resemblance to the intra-block relationship $vol$ in the NCut method\cite{shi2000normalized}, yet there are notable differences in their specifics. In NCut, the intra-block relationship is defined as the sum of the degrees of nodes within a connected block; this degree accounts for the edges that connect each node within the block. Consequently, edges connecting nodes within the block to those outside it are also included in this count. In contrast, while we similarly account for edges within a cluster, we deliberately exclude any edges that connect nodes inside the cluster to those outside it, which provides a more intuitive representation. The distinction between Def. (\ref{def:vol}) and (\ref{R_intra}) becomes evident when examining their respective summation ranges.

Inter-class relationships are also different from the cuts. To ensure the stability of our results, we introduces the margin sets $\c M$ as the partition between clusters. For nodes belonging to different clusters to be connected, they must traverse at least one margin node. This requirement distinguishes Def. (\ref{def:cut}) from (\ref{R_inter}). During the initial stages of the merging process, there exists a subset of nodes that cannot be distinctly assigned to either cluster; we limit their influence on inter-cluster relationships. This strategy effectively mitigates issues such as false pseudo-labels.

%Experiment
\section{Experimental Results and Analysis} \label{sec:Experiment}

\noindent In this section, we present experiments designed to demonstrate the efficacy of our approach, DPSM, and the correctness of our conclusions.

%To count the run time, all experiments are performed on a Windows 10 computer with a 3.60GHz Intel(R) Core(TM) i7-7700 CPU and 32.0 GB RAM, python 3.9.

\begin{table}[h!]
\centering
\caption{The Description of Datasets} \label{datasets}
\begin{center}
\begin{tabular}{ccccc}
\hline
\t{Dataset} & \t{Size} & \t{ Classes} & \t{Dimensions} \\ 
\hline
Circledata	&1,500	&2	&2\\
Jain		&1,000	&2	&2\\
Dermatology&366		&6	&34 \\
Control 	&600		&6	&60\\
Coil20	&1440		&20	&1024\\
Coil100	&7200		&100	&12288\\
MSRA25	&1799		&12	&256\\ 
USPS		&9298		&10	&256\\ \hline
\end{tabular}
\end{center}
\end{table}

\subsection{Datasets and Metrics}

\noindent  We utilize both artificial and real-world datasets to evaluate the proposed method alongside other comparative approaches. The artificial datasets comprise concentric circles (circledata) and two-moon dataset (Jain), which are used to evaluate the methods' ability to manage  irregularly shaped clusters. The real-world datasets include Control\cite{synthetic_control_chart_time_series_139}, Dermatology\cite{dermatology_33}, MSRA25\cite{zhang2012finding}, Coil20\cite{nene1996columbia}, Coil100\cite{Nene1996COIL100}, and USPS\cite{Hull1994ADF}. Detailed information regarding all datasets is presented in Table \ref{datasets}. These datasets have all been reduced to two dimensions utilizing the t-SNE method\cite{van2008visualizing} to enhance clustering and visualization.

V-measure (VM)\cite{rosenberg2007v}, Adjusted Rand Index (ARI)\cite{hubert1985comparing}, and Adjusted Mutual Information (AMI)\cite{vinh2009information} are three widely used metrics for evaluating clustering results. VM considers both homogeneity and completeness; ARI focuses on the consistency of node pairs across different states; AMI quantifies the similarity between two sets of labels through mutual information. We employ these three metrics to assess the performance of the algorithms.

Notably, some algorithms require the feature space as input, while others depend on a graph or inter-node similarities. To achieve unification in our analysis, we apply the $k$-nearest neighbor method to nodes within the feature space, resulting in an undirected graph after symmetric  normalization\cite{mcinnes2018umap}, as illustrated below:
\eqta\label{build graph}
\alna
& a(i,j) = \exp(-\sigma||\b x_i - \b x_j||_2), j \in \r{kNN}(i), \\
& \tilde a(i,j) = 
\left\{ \alna
& \frac{a(i,j)}{\sum_{j \in \r{kNN}(i)} a(i,j)} ,& j \in \r{kNN}(i),\\
& 0 ,& \r{ otherwise},
\alnb \right.\\
& w(i,j) = \tilde a(i,j) + \tilde a(j,i) - \tilde a(i,j) \cdot \tilde a(j,i),
\alnb
\eqtb
where $\r{kNN}(i)$ denotes the $k$-nearest neighbors of node $i$. The parameters employed include a nearest neighbor value of $k=20$, and the kernel parameter $\sigma$ is $0.1 d_{max}$ where $d_{max}$ represents the maximum distance between any two nodes in the dataset.

\begin{table*}[t!]
\centering
\caption{Clustering Results With No Specified Number of Classes} \label{T1}
\begin{center}
\begin{tabular*}{\textwidth}{@{\extracolsep{\fill}} l c c c c c c c c @{}}
\hline
\t{Algorithm}  & \t{VM} & \t{ ARI} & \t{ AMI} & { \textit{cluster}} & \t{VM} & \t{ ARI} & \t{ AMI} & {\textit{cluster}}\\ \hline
&\textit{Circledata}& & & 2 &\textit{Jain} & & & 2\\
\t{DPSM(Ours)}  &	\t{1.0000}&	\t{1.0000}&	\t{1.0000}&	2 &	\t{1.0000}&	\t{1.0000}&	\t{1.0000}&	2\\
DPC   &	0.4268&	0.2393&	0.4255&	6 &	0.6802&	0.5401&	0.6797&	4\\
MeanShift &	0.4714&	0.2228&	0.4700&	8 &	0.5260&	0.3525&	0.5250&	6\\
DBSCAN&	0.0962&	-0.0698&	0.0406&	88 &	0.1183&	0.0014&	0.0657&	72\\
Affinity Propagation&	0.3037&	0.0464&	0.2998&	36&	0.3545&	0.0892&	0.3512&	18\\
LPA   &	0.3019&	0.0483&	0.2962&	33 &	0.3619&	0.0945&	0.3577&	26\\ \hline
&\textit{Dermatology}& & & 6 &\textit{Control} & & & 6\\
\t{DPSM(Ours)}  &	\t{0.9253}&	\t{0.8629}&	\t{0.9239}&	5&	\t{0.8520}&	\t{0.6824}&	\t{0.8507}&	4\\
DPC   &	0.7465&	0.5727&	0.7438&	3&	0.7602&	0.5694&	0.7588&	3\\
MeanShift &	\t{0.9253}&	\t{0.8629}&	\t{0.9239}&	5&	0.7878&	0.6155&	0.7861&	4\\
DBSCAN&	0.0779&	-0.0049&	0.0229&	10&	0.0650&	0.0003&	0.0250&	11\\
Affinity Propagation&0.6964&	0.3490&	0.6780&	18&	0.6885&	0.3252&	0.6731&	27\\
LPA   &	0.7885&	0.6063&	0.7804&	11&	0.7215&	0.4337&	0.7121&	18\\ \hline
&\textit{Coil20}& & & 20 &\textit{Coil100} & & & 100\\
\t{DPSM(Ours)}  &	\t{0.8080}&	\t{0.5892}&	\t{0.7975}&	22&	\t{0.8831}&	\t{0.6630}&	\t{0.8590}&	111\\
DPC   &	0.7099&	0.3493&	0.7048&	7&	0.4121&	0.0408&	0.4078&	4\\
MeanShift &	0.7048&	0.3693&	0.6982&	9&	0.4795&	0.0703&	0.4736&	6\\
DBSCAN&	0.0110&	0.0000&	0.0025&	5&	0.1154&	0.0002&	0.0221&	218\\
Affinity Propagation&	0.7524&	0.3027&	0.6969&	95&	0.8575&	0.4311&	0.7885&	423\\
LPA   &	0.8037&	0.5434&	0.7819&	45&	0.8721&	0.5414&	0.8270&	224\\ \hline
&\textit{MSRA25}& & & 12 &\textit{USPS} & & & 10\\
\t{DPSM(Ours)}  &	\t{0.7934}&	\t{0.5340}&	\t{0.7896}&	16&	\t{0.8731}&	\t{0.8568}&	\t{0.8728}&	9\\
DPC   &	0.3564&	0.1380&	0.3541&	3&	0.6012&	0.2749&	0.6009&	4\\
MeanShift &	0.6192&	0.3670&	0.6150&	9&	0.7396&	0.6801&	0.7392&	9\\
DBSCAN&	0.1923&	0.0006&	0.0629&	117&	0.1366&	0.0021&	0.0437&	428\\
Affinity Propagation&	0.7179&	0.2449&	0.6925&	81&	0.5177&	0.0358&	0.4870&	452\\
LPA   &	0.7657&	0.3815&	0.7520&	52&	0.5444&	0.0640&	0.5270&	268\\ \hline
\end{tabular*}
\end{center}
\end{table*}

\subsection{Clustering Experiments With Unknown Number of Classes} \label{ex:unknown}

\noindent Among the methods that do not require a predetermined number of classes, or can adaptively determine the number of classes, we select five representative methods for comparison:

1) \t{DPC}\cite{rodriguez2014clustering}: DPC is introduced in Section \ref{DPC}.For parameters, the distance measure employs Euclidean distance after dimensionality reduction, and the value of $D_c$ is determined using an empirical discriminant method, defined as the 2\% position after sorting all distances in ascending order. The number of classes is not specified; therefore, the determination of whether a node serves as a cluster center relies on the thresholds of $\rho$ and $\delta$, which represent the mean values of the maximum and minimum, respectively.

2) \t{Meanshift}\cite{fukunaga1975estimation}: MeanShift serves as an iterative localization method operating within a non-parametric feature space. The bandwidth parameter is initially estimated using the histogram, followed by an evaluation of the interval ranging from half to twice this estimate in order to determine the optimal parameter.

3) \t{DBSCAN}\cite{ester1996density}: DBSCAN is a density-based clustering algorithm that relies on two primary parameters, $\epsilon$ and $min\_samples$. To identify the optimal parameter values, the Grid Search strategy\cite{bishop2006pattern} is employed. The probing range for $\epsilon$ spans from $0.1$ to $0.5$, while the range for of $min\_samples$ spans from $2$ to $20$.

4) \t{Affinity Propagation}\cite{frey2007clustering}: The Affinity Propagation method determines the similarity between ndoes by exchanging messages and selects "exemplars" as cluster centers. The Grid Search strategy is also employed: The probe range of $damping$ is from $0.5$ to $0.9$, while that of $preference$ is from $-100$ to $0$.

5) \t{LPA}\cite{raghavan2007near}: LPA, introduced in Section \ref{Propagation}, can also be employed as an unsupervised community detection method that identifies community structures through label propagation among nodes. The graph it uses is constructed based on the weight function $w(i,j)$, as described in Eq. (\ref{build graph}).

In DPSM, the number of iterations of density propagation is $100$, and the termination for merging occurs when the measure $CluCut$ of the subsequent merge falls below half of the preceding one, which is discussed in Section \ref{termination} in detail. During the merging process, in order to exclude outliers, we also eliminate clusters that have both a node count and intra-class relationship less than 5\% of the average.

Since several methods are random-dependent, they were executed five times with seeds from 0 to 4, and the average metric is calculated. 

The results are presented in TABLE \ref{T1}. In addition to the three previously mentioned metrics, the $cluster$ column indicates the number of clusters identified by the method. The actual count is documented on the line corresponding to the italicized names of the datasets.
For the majority of the datasets, dPSM identified a number that closely approximates the true count of clusters, resulting in superior outcomes. While Meanshift and other techniques may estimate the actual number of clusters for certain datasets, there exists a significant discrepancy between these identified clusters and the real classes, leading to suboptimal performance. Methods such as LPA often yield a greater number of clusters than what is realistically present; however, they still achieve VM and AMI scores in some datasets that are second only to those produced by DPSM. This suggests that the smaller clusters detected by these methods composing the true classes and require further merging.

\begin{table*}[t!]
\centering
\caption{Clustering Results With a Defined Number of Classes} \label{T2}
\begin{center}
\begin{tabular*}{\textwidth}{@{\extracolsep{\fill}} l c c c c c c c c @{}}
\hline
\t{Algorithm}  & \t{VM} & \t{ ARI} & \t{ AMI} & { \textit{cluster}} & \t{VM} & \t{ ARI} & \t{ AMI} & { \textit{cluster}}\\ \hline
&\textit{Circledata}& & & 2 &\textit{Jain} & & & 2\\
\t{DPSM(Ours)}  &	\t{1.0000}&	\t{1.0000}&	\t{1.0000}&	2&	\t{1.0000}&	\t{1.0000}&	\t{1.0000}&	2 \\
DPC   &	0.3977&	0.3197&	0.3973&	2&	0.4308&	0.3740&	0.4303	&2\\
Spectral Clustering &	\t{1.0000}&	\t{1.0000}&	\t{1.0000}&	2&	\t{1.0000}&	\t{1.0000}&	\t{1.0000}&	2 \\
K-Means&	0.1866&	0.2105&	0.1860&	2&	0.2155&	0.2823&	0.2149	&2\\
Birch &	0.2828&	0.1176&	0.2822&	2&	0.4461&	0.3964&	0.4457	&2\\
Agglomerative Clustering&	0.4111&	0.3428&	0.4106&	2&	0.4426&	0.3914&	0.4422	&2\\ \hline
&\textit{Dermatology}& & & 6 &\textit{Control} & & & 6\\
\t{DPSM(Ours)}  &	0.9272&	0.9308&	0.9256&	6 &	0.8036&	0.6492&	0.8011&	6\\
DPC   &	\t{0.9298}&	\t{0.9347}&	\t{0.9283}&	6&	0.7942&	0.6560&	0.7916&	6\\
Spectral Clustering &	0.8927&	0.7963&	0.8903&	6&	0.7300&	0.5424&	0.7266&	6\\
K-Means&	0.8474&	0.7942&	0.8441&	6&	0.7640&	0.6630&	0.7611&	6\\
Birch &	0.9220&	0.9271&	0.9203&	6&	0.7473&	0.5913&	0.7442&	6\\
Agglomerative Clustering&	0.9294&	0.9307&	0.9279&	6&	\t{0.8266}&	\t{0.7210}&	\t{0.8245}&	6\\ \hline
&\textit{Coil20}& & & 20 &\textit{Coil100} & & & 100\\
\t{DPSM(Ours)}  &	\t{0.8828}&	0.7082&	\t{0.8771}&	20&	\t{0.9189}&	0.7315&	\t{0.9038}&	100\\
DPC   &	0.8704&	0.6618&	0.8639&	20 &	0.9044&	0.6195&	0.8886&	100\\
Spectral Clustering &	0.7176&	0.2893&	0.7024&	20&	0.9112&	0.6658&	0.8949&	100\\
K-Means&	0.8396&	0.7095&	0.8320&	20&	0.9030&	0.7324&	0.8843&	100\\
Birch &	0.8523&	0.7113&	0.8452&	20&	0.9136&	\t{0.7481}&	0.8971&	100\\
Agglomerative Clustering&	0.8598&	\t{0.7264}&	0.8531&	20&	0.9142&	0.7499&	0.8978&	100\\ \hline
&\textit{MSRA25}& & & 12 &\textit{USPS} & & & 10\\
\t{DPSM(Ours)}  &	\t{0.7934}&	\t{0.5340}&	\t{0.7896}&	16&	\t{0.8902}&	\t{0.9079}&	\t{0.8900}&	10\\
DPC   &	0.7430&	0.4765&	0.7392&	12&	0.8496&	0.7805&	0.8493&	10\\
Spectral Clustering &	0.6195&	0.2610&	0.6133&	12&	0.8257&	0.7488&	0.8254&	10\\
K-Means&	0.7386&	0.5129&	0.7349&	12&	0.8320&	0.7700&	0.8317&	10\\
Birch &	0.7520&	0.5052&	0.7484&	12&	0.8414&	0.7683&	0.8411&	10\\
Agglomerative Clustering&	0.7603&	0.5250&	0.7569&	12&	0.8297&	0.7474&	0.8294&	10\\ \hline
\end{tabular*}
\end{center}
\end{table*}

\subsection{Clustering Experiments With a Defined Number of Classes}

\noindent Numerous methods necessitate or leverage a predetermined number of classes to achieve optimal effectiveness. We have selected five such methods:

1) \t{DPC}\cite{rodriguez2014clustering}: The DPC method is employed once more; however, this time it operates with a specified number of target clusters.

2) \t{Spectral Clustering}\cite{ng2001spectral}: This method employs the NCut introduced in Section \ref{Spectral Clustering}. To ensure the consistency of the graph, a weight function $w(i,j)$ is employed, as elaborated in Eq. (\ref{build graph}).

3) \t{$\b{k}$-Means}\cite{lloyd1982least}: The initial approach employed is $k$-Means++\cite{arthur2006k}, and the method has an iteration upper bound of 300.

4) \t{Birch}\cite{zhang1996birch}: Birch builds a characteristic feature tree to incrementally and dynamically cluster incoming nodes. In this tree, each node has no more than 50 branches, and the merging threshold is set at 0.5.

5) \t{Agglomerative Clustering}\cite{murtagh2014ward}: In this hierarchical clustering method, the inter-cluster similarity is measured using Euclidean distance, and the merging processes rely on Ward Linkage.

The settings for DPSM and DPC that are not explicitly mentioned align with those used in the experiments presented in TABLE \ref{T1}. We continue to utilize the same 8 datasets and 3 metrics for evaluating the algorithms. Additionally, the random-dependent methods were executed five times using random seeds ranging from 0 to 4 to obtain average metrics.

As demonstrated in Table \ref{T2}, DPSM consistently achieves either the best or second-best performance. As methods capable of either specifying the number of clusters or autonomously determining it, both DPSM and DPC show improvements or maintain stability in clustering results after the number of clusters is established. For the MSRA25 dataset, the data is automatically partitioned into 16 connected blocks, with no connecting edges between any two blocks. DPSM cannot proceed to merge further after identifying these 16 connected blocks; therefore, the number of clusters reported in the table is 16 rather than the actual count of 12. In this case, DPSM can still obtain the best clustering results. Spectral clustering aligns with the idea of merging process in DPSM; however, it does not initiate with a density-based partitioning. Consequently, this approach tends to perform well on more structured datasets, such as the artificial datasets Circledata and Jain, while exhibiting suboptimal performance on real-world datasets.

\begin{figure*}[t!]
\centering
\begin{tabular}{@{}*{6}{c}@{}} % 6列居中对齐
    \includegraphics[width=0.15\textwidth]{figures/Ours\_density.pdf} &
    \includegraphics[width=0.15\textwidth]{figures/Ours1\_density.pdf} &
    \includegraphics[width=0.15\textwidth]{figures/Ours2\_density.pdf} &
    \includegraphics[width=0.15\textwidth]{figures/Ours3\_density.pdf} &
    \includegraphics[width=0.15\textwidth]{figures/Ours4\_density.pdf} &
    \includegraphics[width=0.15\textwidth]{figures/Ours5\_density.pdf} \\
    
    \includegraphics[width=0.15\textwidth]{figures/Ours\_catgory.pdf} &
    \includegraphics[width=0.15\textwidth]{figures/Ours1\_catgory.pdf} &
    \includegraphics[width=0.15\textwidth]{figures/Ours2\_catgory.pdf} &
    \includegraphics[width=0.15\textwidth]{figures/Ours3\_catgory.pdf} &
    \includegraphics[width=0.15\textwidth]{figures/Ours4\_catgory.pdf} &
    \includegraphics[width=0.15\textwidth]{figures/Ours5\_catgory.pdf} \\
    
    \includegraphics[width=0.15\textwidth]{figures/Ours\_initial.pdf} &
    \includegraphics[width=0.15\textwidth]{figures/Ours1\_initial.pdf} &
    \includegraphics[width=0.15\textwidth]{figures/Ours2\_initial.pdf} &
    \includegraphics[width=0.15\textwidth]{figures/Ours3\_initial.pdf} &
    \includegraphics[width=0.15\textwidth]{figures/Ours4\_initial.pdf} &
    \includegraphics[width=0.15\textwidth]{figures/Ours5\_initial.pdf} \\

    \includegraphics[width=0.15\textwidth]{figures/Ours\_end.pdf} &
    \includegraphics[width=0.15\textwidth]{figures/Ours1\_end.pdf} &
    \includegraphics[width=0.15\textwidth]{figures/Ours2\_end.pdf} &
    \includegraphics[width=0.15\textwidth]{figures/Ours3\_end.pdf} &
    \includegraphics[width=0.15\textwidth]{figures/Ours4\_end.pdf} &
    \includegraphics[width=0.15\textwidth]{figures/Ours5\_end.pdf} \\

	With Density & Random-1 & Random-2 & Random-3 & Random-4 & Random-5
\end{tabular}
\caption{Experiments on the Necessity of Propagated Density (the USPS dataset)\\
The images in the first and second rows illustrate the densities and categories of nodes, whereas those in the third and fourth rows depict the initial and final outcomes of clustering.}
\label{Fig:Density}
\end{figure*}

\begin{table}[h]
\centering
\caption{Experiments on the Necessity of Propagated Density} \label{T3}
\begin{center}
\begin{tabular}{cccccc}
\hline
\t{Algorithm}  & \t{VM} & \t{ ARI} & \t{ AMI} & { \textit{cluster}} \\ \hline		
&\textit{Coil100}& & & 100\\
With Density  &0.8831&	\t{0.6630}&	\t{0.8590}&	111\\
Random-1&0.8691&	0.4638&	0.8534&	61\\
Random-2&	0.8612&	0.4505&	0.8452&	58\\
Random-3&\t{0.8884}&	0.6406&	0.8586&	158\\
Random-4& 0.8507&	0.4057&	0.8344&	55\\
Random-5& 0.8852&	0.6135&	0.8505&	186\\
Random-Average&	0.8709&	0.5148&	0.8484&	103.6\\ \hline
&\textit{USPS}& & & 10\\
With Density  &\t{0.8731}&	\t{0.8568}&	\t{0.8728}&	9\\
Random-1 &0.7781&	0.6397&	0.7779&	6\\ 
Random-2 &0.8368&	0.7855&	0.8366&	8\\ 
Random-3 &0.7947&	0.6624&	0.7944&	7\\ 
Random-4 & 0.8509&	0.8192&	0.8506&	10\\
Random-5&	0.7790&	0.6174&	0.7787&	6\\
Random-Average&	0.8154&	0.7048&	0.8076&	7.4\\  \hline
\end{tabular}
\end{center}
\end{table}

\subsection{Ablation Experiments on the Necessity of Propagated Density}

\noindent Through experiments, we demonstrate that the density determined by propagation is highly valuable for clustering purposes. This assertion is substantiated through a comparison with random density. The density established via propagation primarily relies on identifying neighborhood relationships, making it mismatch to compare with other density determination methods that require global distance calculations, such as those employed in DPC.

In smaller datasets, the difference is not statistically significant. Therefore, we emphasize the distinction in performance when employing propagation density on the Coil100 and USPS data sets, as illustrated in Table \ref{T3}. In addition to the standard application of DPSM (referred to as 'with density' in the table), we also compute metrics based on randomly assigned densities.  After obtaining densities, we partition and merge nodes and clusters to obtain the final outcome.  The results derived from five random assignments are labeled as 'Random-1' to 'Random-5', while their average metrics are denoted as 'Random-Average'. The method of randomly assigning density may yield a clustering effect that is comparable to that achieved through the use of propagation density; however, the results are not stable.

In Fig. \ref{Fig:Density}, several visualizations support this assertion:

1) In images in the first row, the color depth of each node corresponds to its assigned density. The process of density propagation results in a significantly higher density for nodes within the class compared to those outside or drifting away. By examining the lighter nodes, one can discern the delineating boundaries between the actual classes.

2) The images in the second row categorize nodes based on their densities, with green indicating identified roots, blue representing internal nodes distinct from roots, and red denoting all other nodes. After applicating propagated density, there is a marked reduction in the number of roots, which are predominantly distributed within their respective classes; this positively contributes to the final clustering outcome. The randomly assigned density may result in certain outliers being perceived as root nodes, thereby disrupting the subsequent merging process.

3) The images presented in the third row illustrate the node partitioning prior to any merging processes.  In this context, gray indicates nodes where category determination remains ambiguous. It is evident that the propagated density not only reduces the number of initial clusters but also decreases the count of undetermined nodes. Notably, these clusters do not intersect with the boundaries of their corresponding real classes and therefore do not diminish homogeneity.

4) Finally, images in the fourth row display clustering results at termination. DPSM successfully avoids misassigning nodes from one class to another—a common issue observed in alternative scenarios. As illustrated in the figures, the two classes located in the upper right corner exhibit a high degree of proximity. Consequently, DPSM is inclined to merge these classes.

\begin{figure*}[t!]
\centering
\begin{tabular}{@{}*{6}{c}@{}} % 6列居中对齐
    \includegraphics[width=0.15\textwidth]{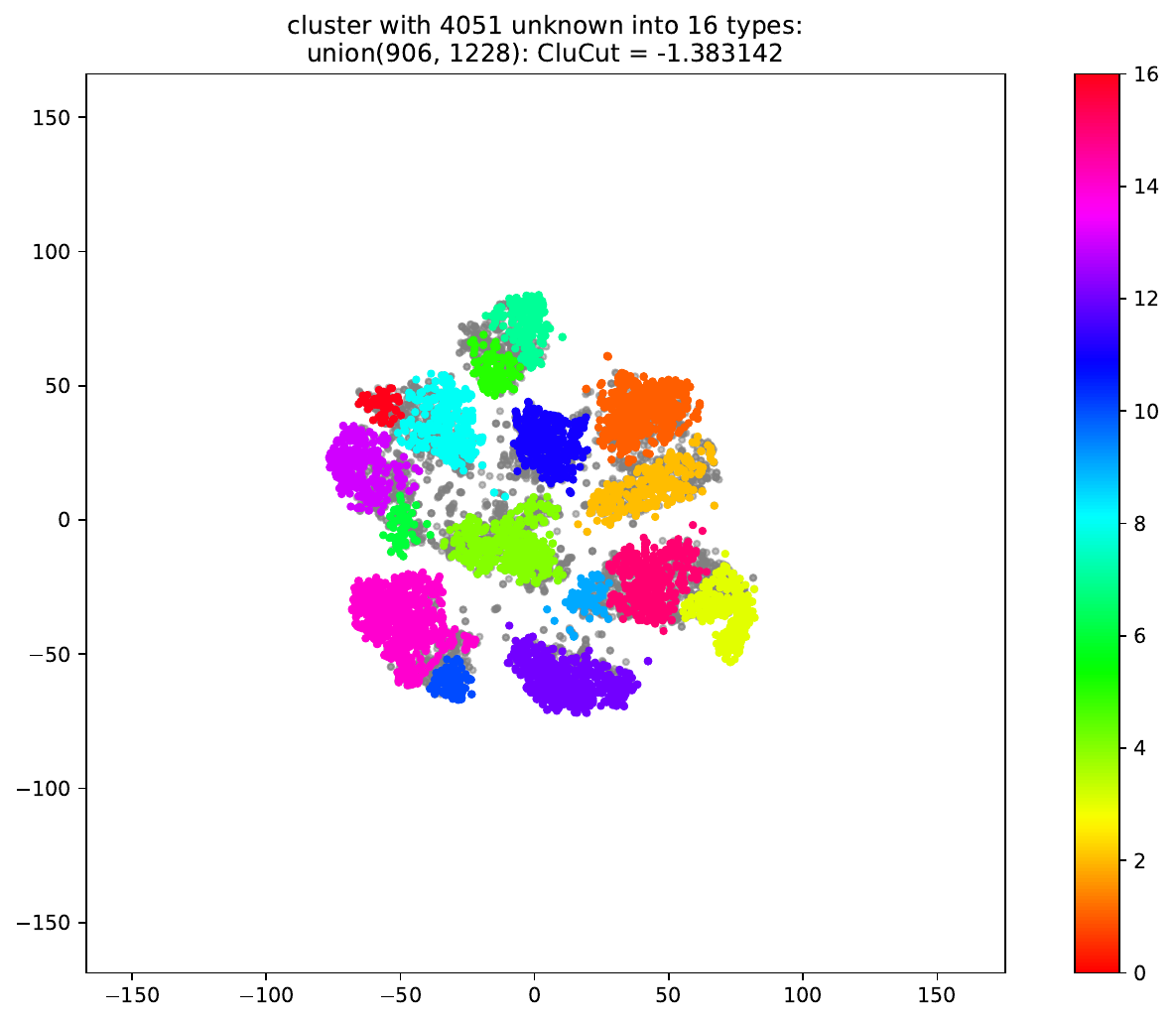} &
    \includegraphics[width=0.15\textwidth]{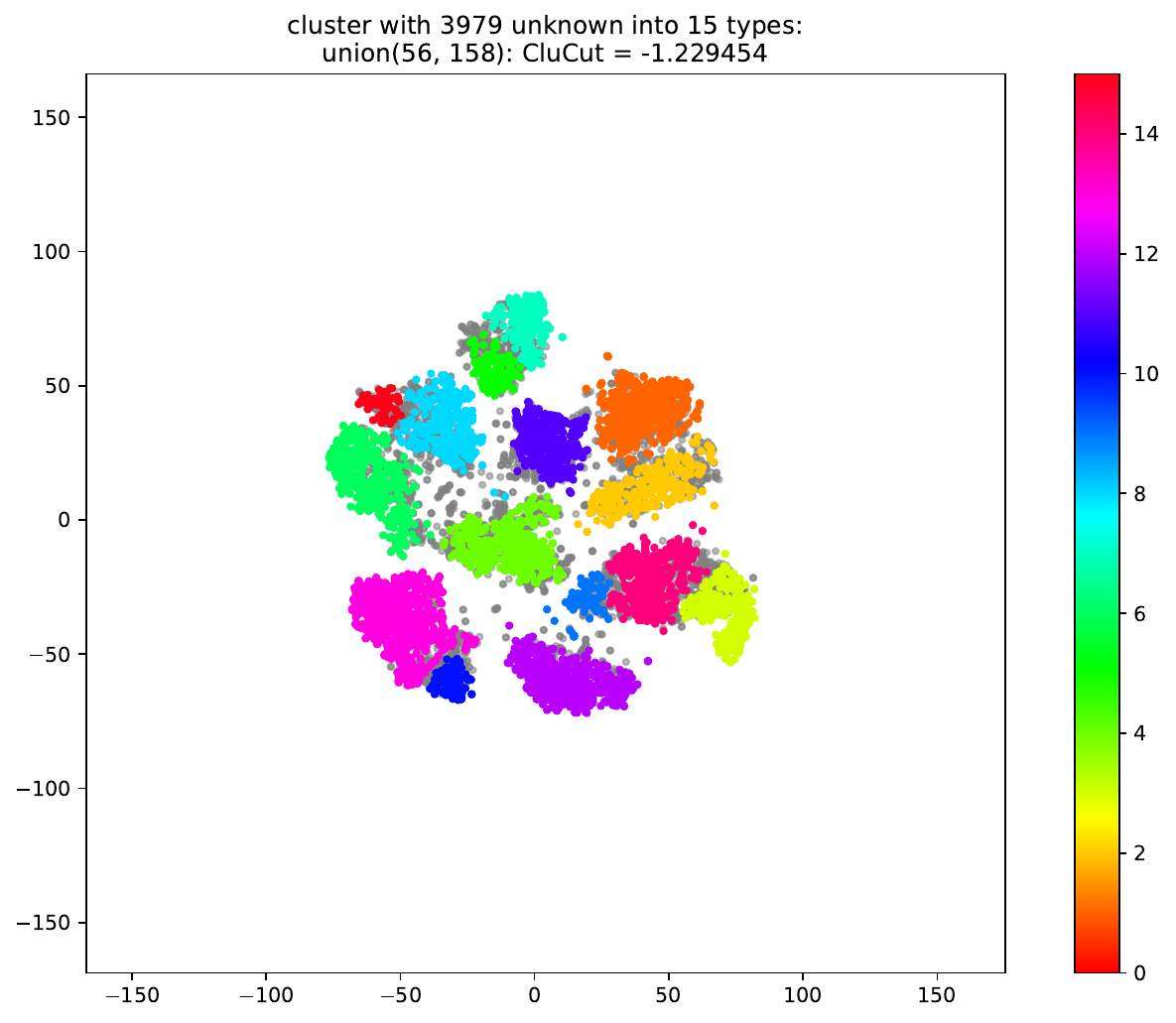} &
    \includegraphics[width=0.15\textwidth]{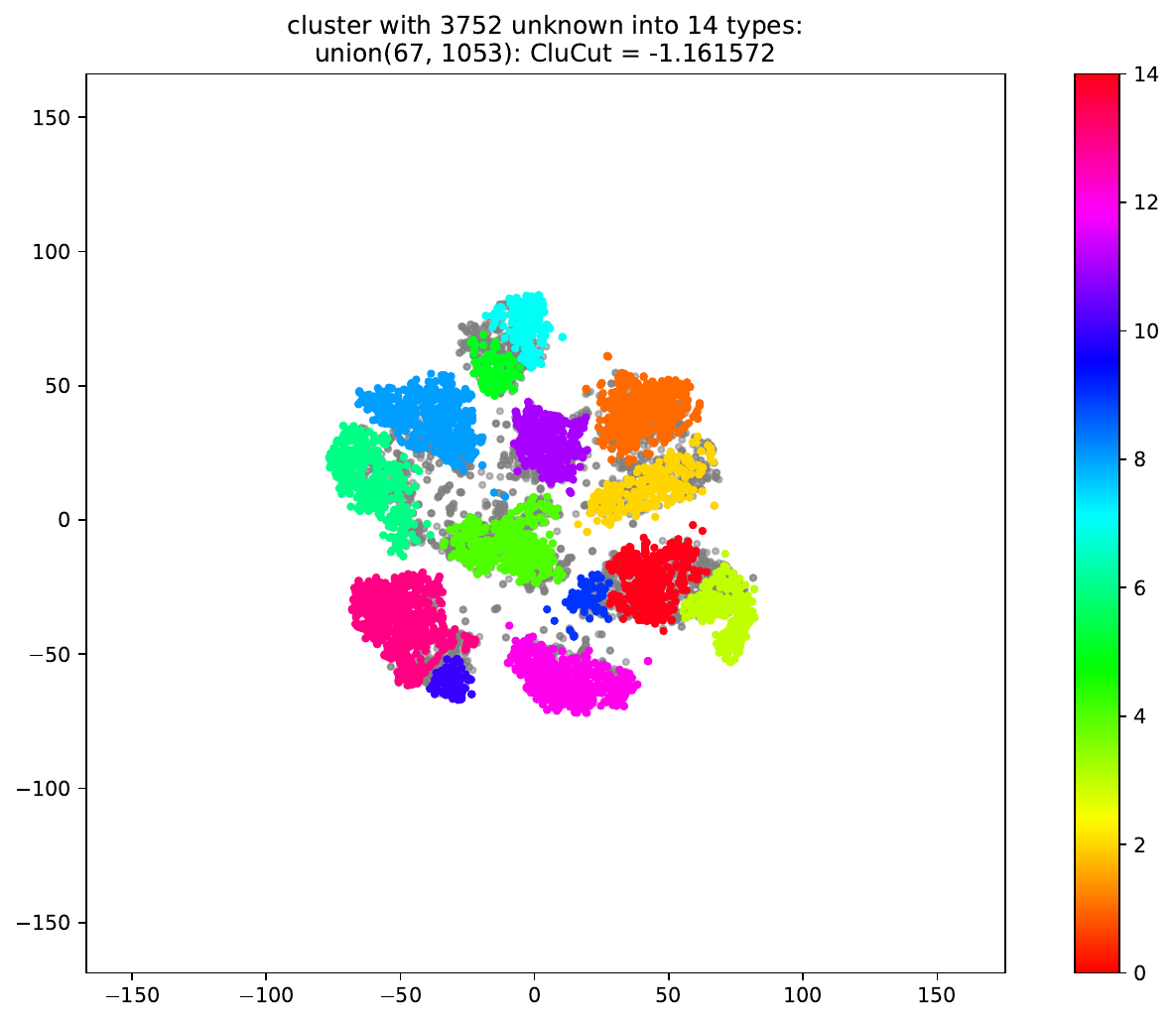} &
    \includegraphics[width=0.15\textwidth]{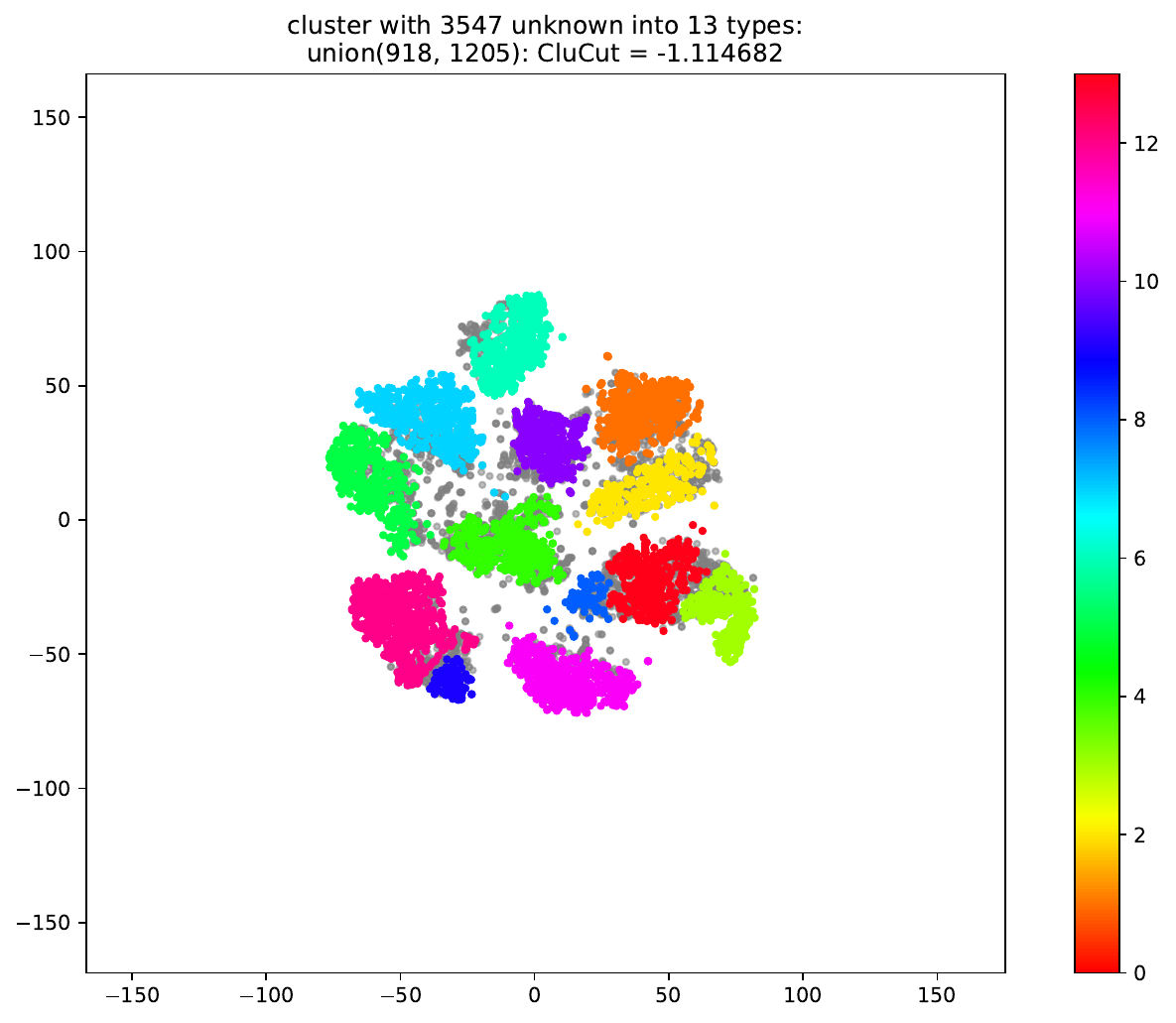} &
    \includegraphics[width=0.15\textwidth]{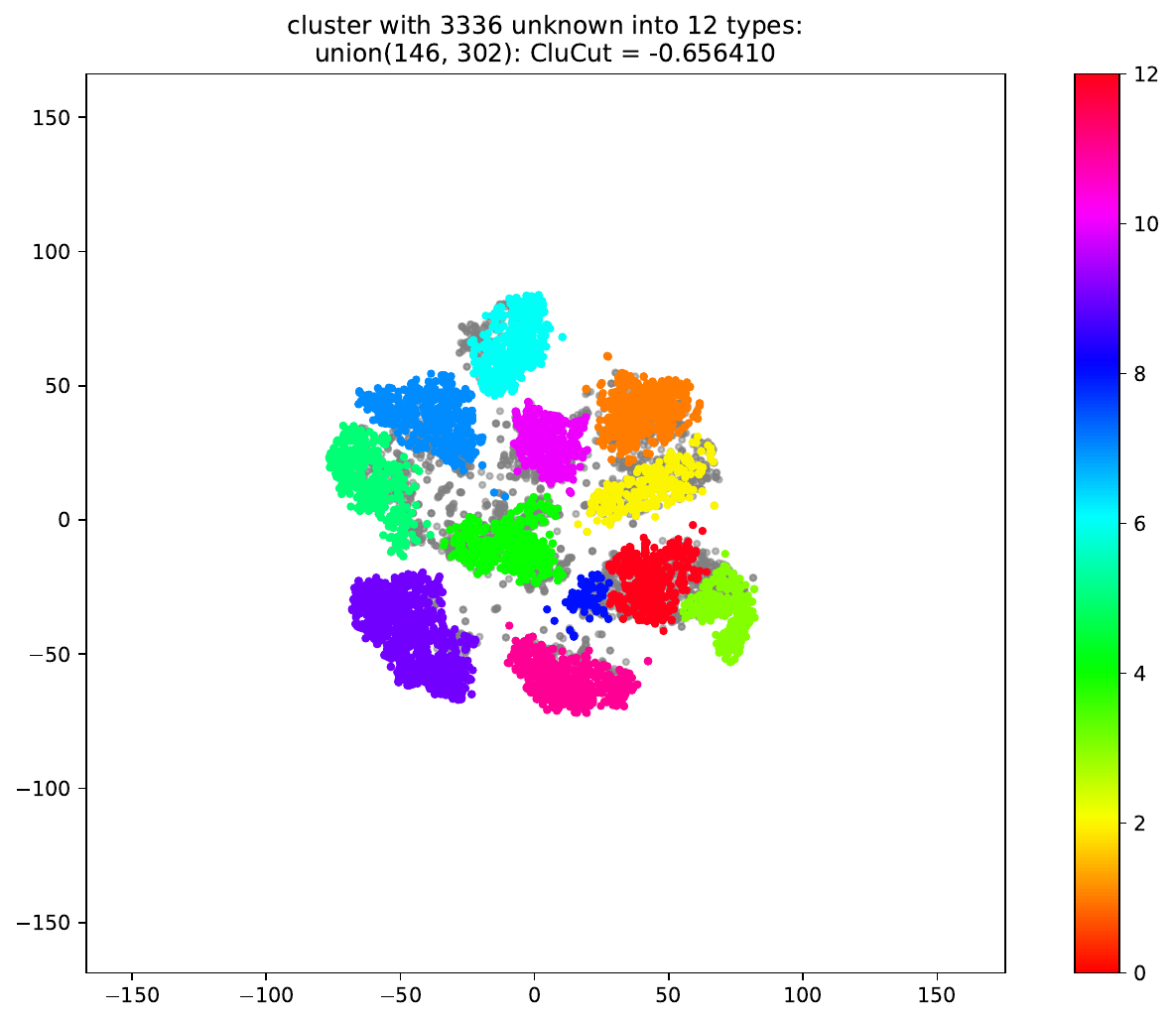} &
    \includegraphics[width=0.15\textwidth]{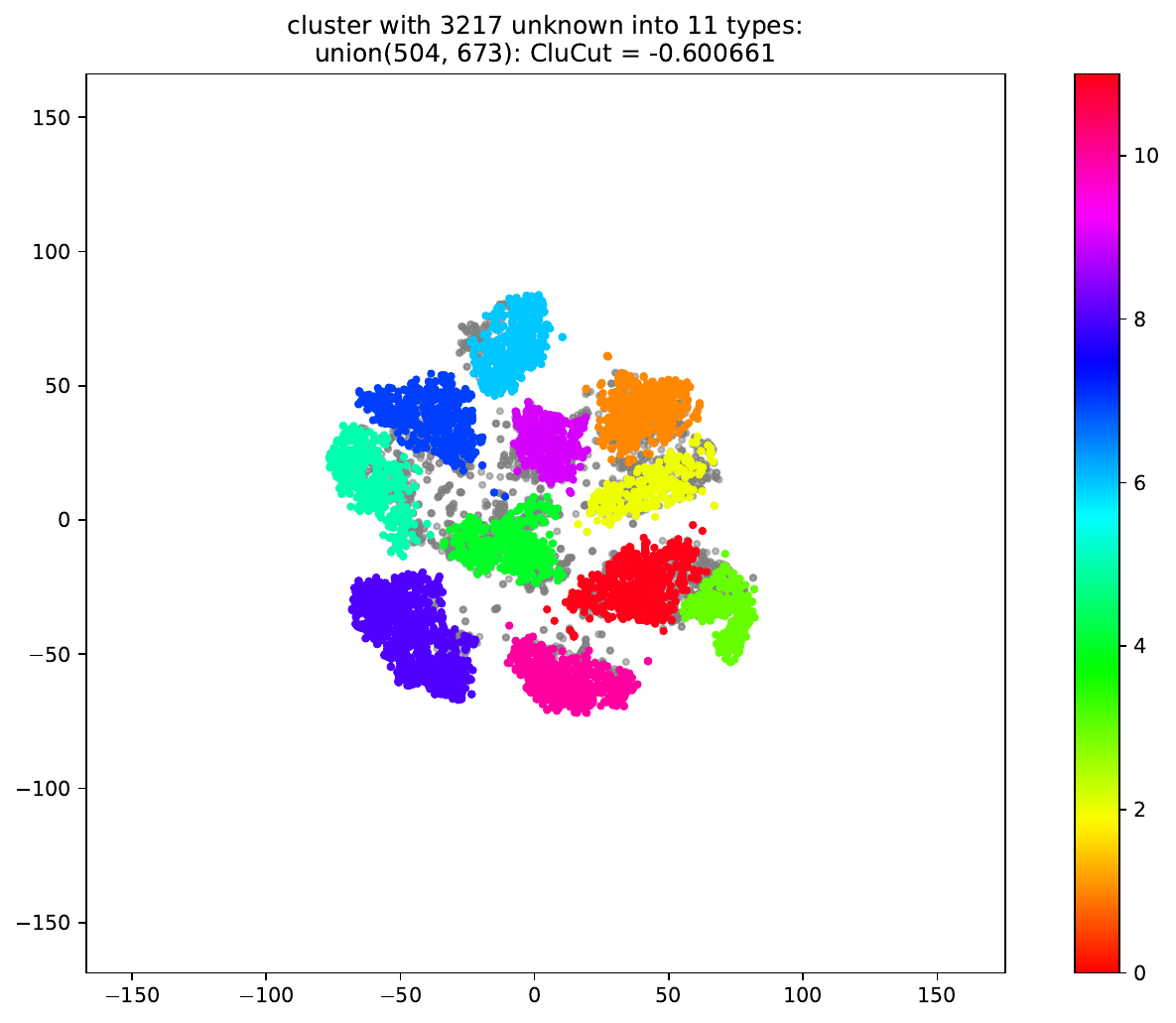} \\
16 clusters& 15 clusters& 14 clusters& 13 clusters&12clusters&11 clusters  \\
    \includegraphics[width=0.15\textwidth]{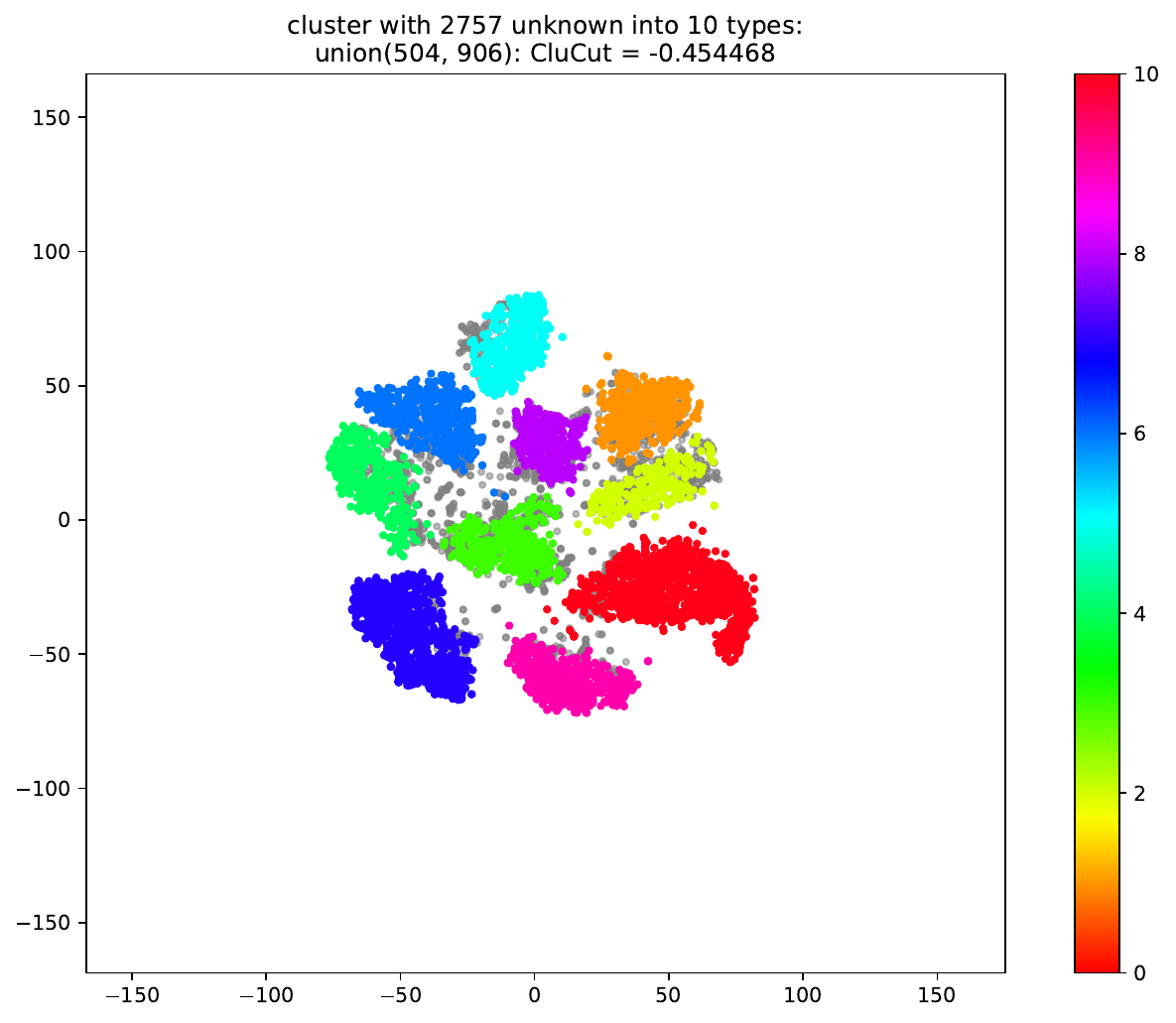} &
    \includegraphics[width=0.15\textwidth]{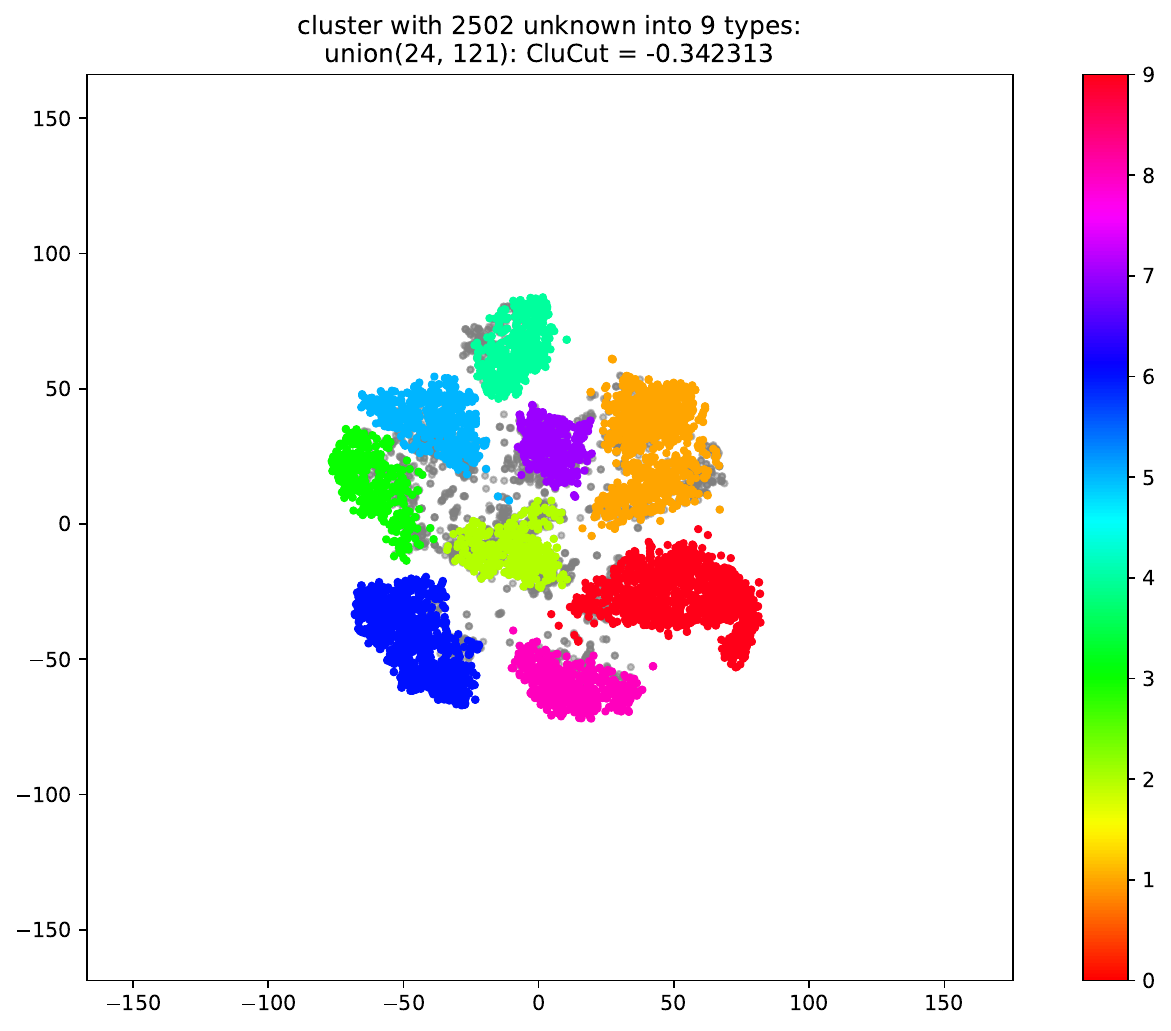} &
    \includegraphics[width=0.15\textwidth]{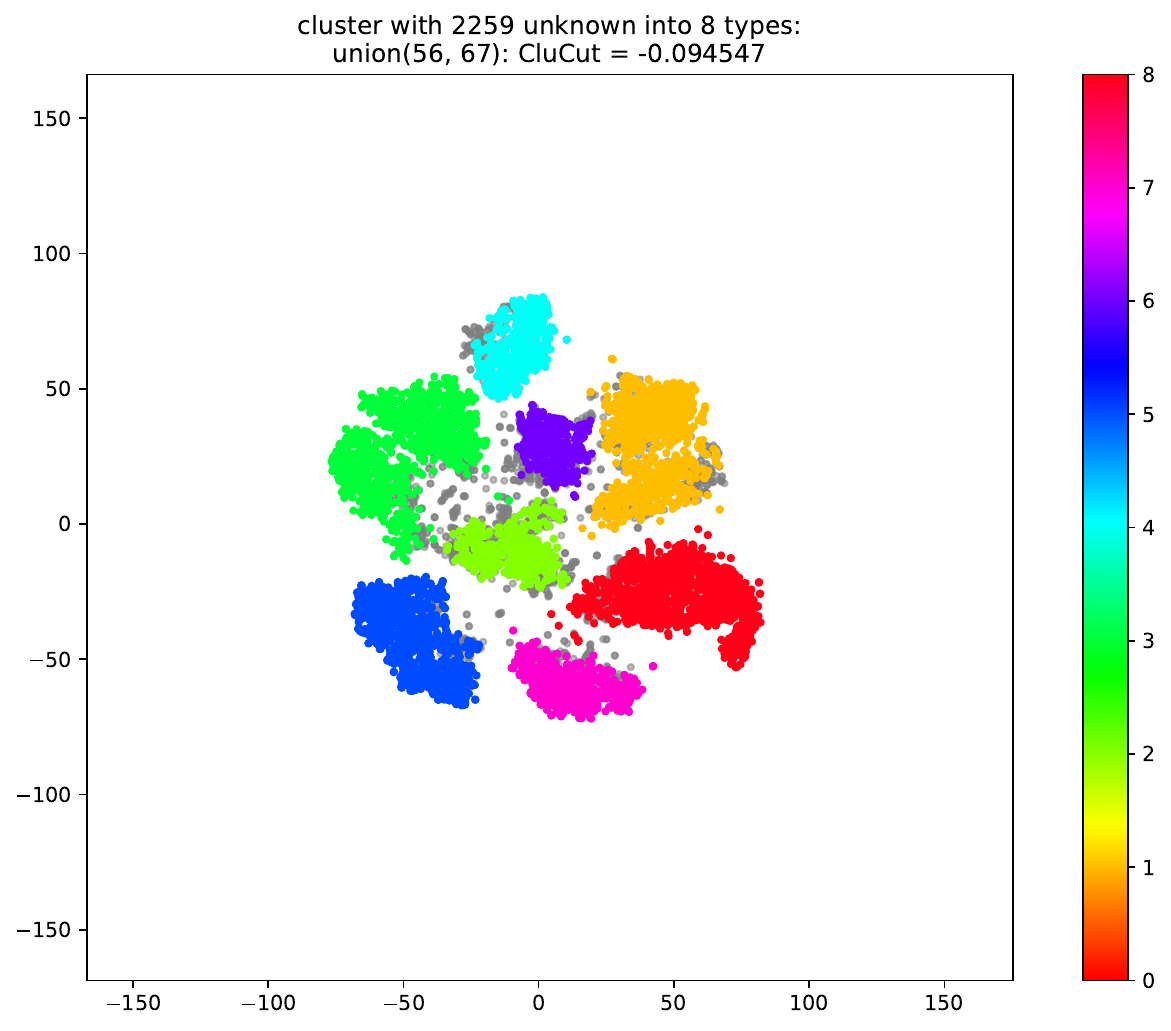} &
    \includegraphics[width=0.15\textwidth]{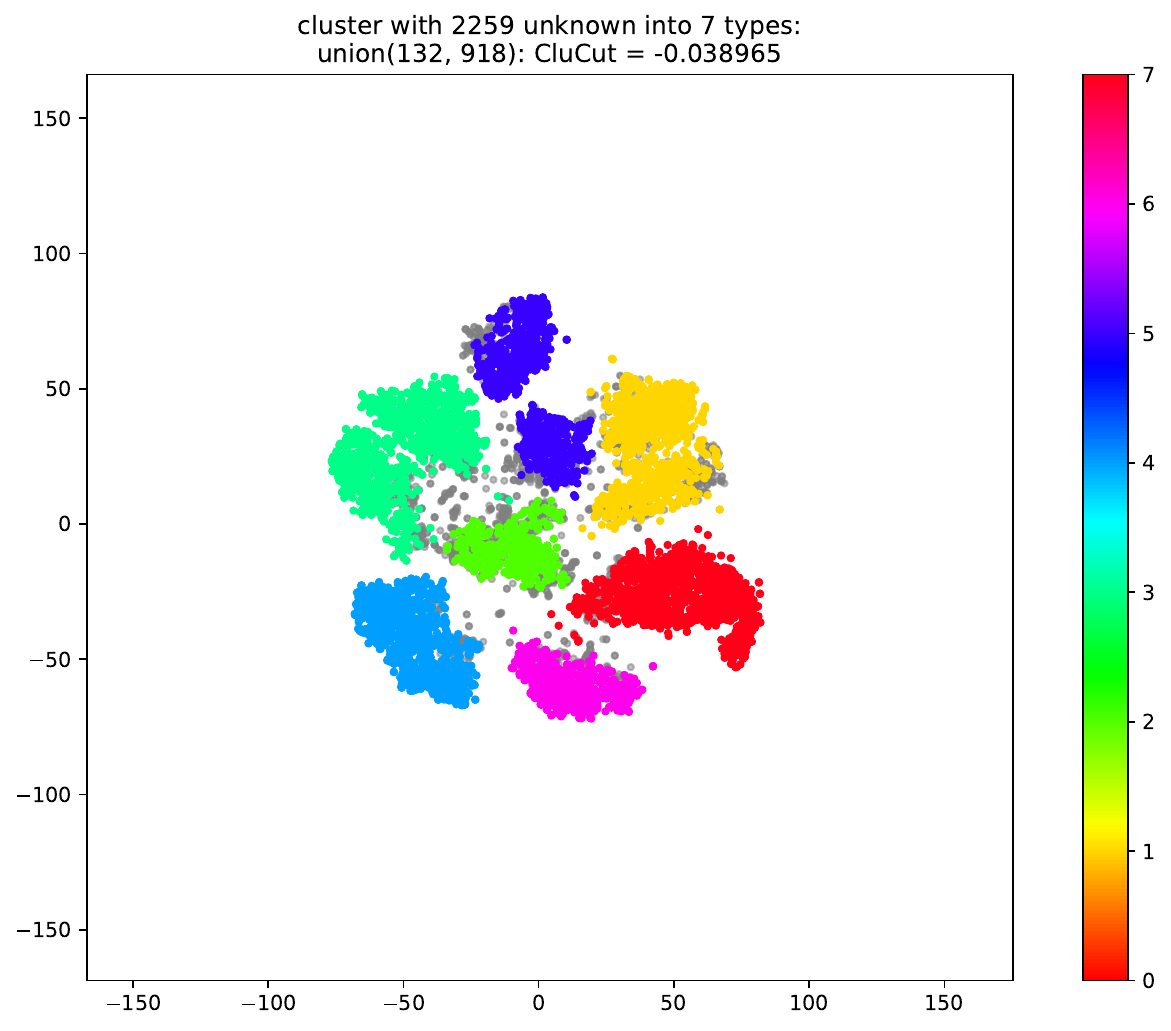} &
    \includegraphics[width=0.15\textwidth]{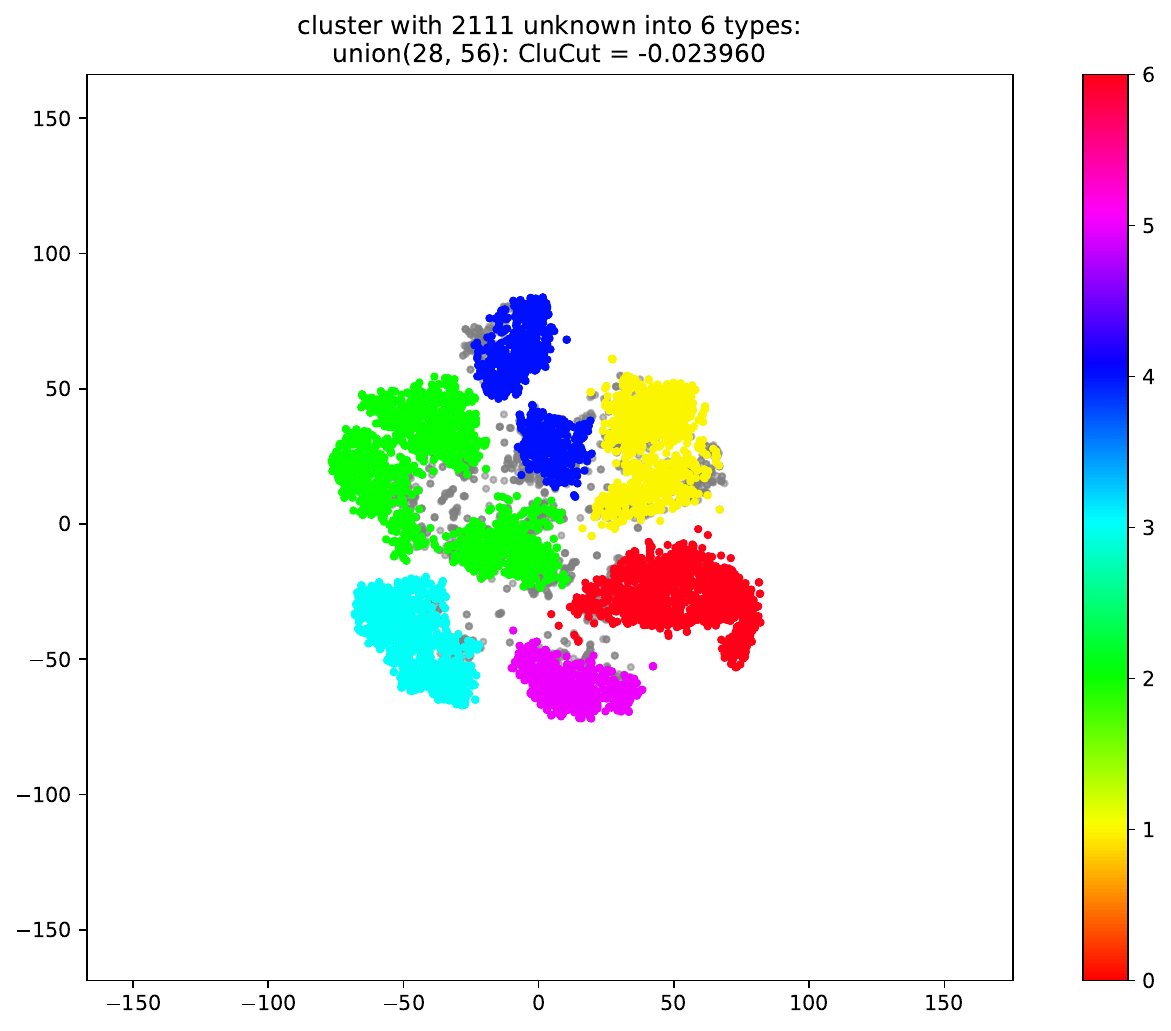} &
    \includegraphics[width=0.15\textwidth]{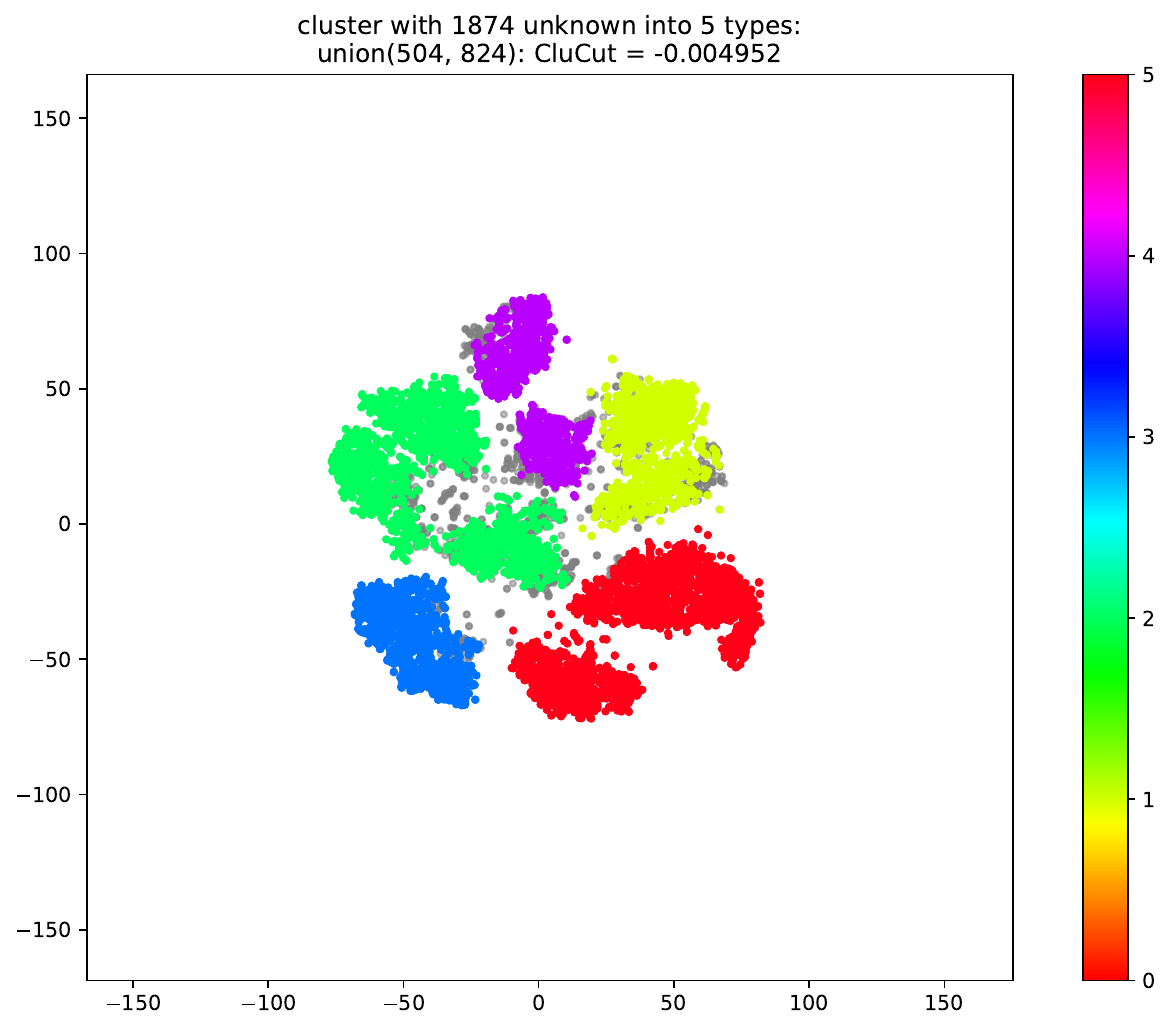} \\
10 clusters& 9 clusters& 8 clusters& 7 clusters&6 clusters&5 clusters  \\
(Actual Number)&(Termination)
\end{tabular}
\caption{Clusters in the Process of Merging (the USPS dataset)\\
The actual number of clusters is 10, DPSM will terminate when there're 9 clusters.}
\label{Fig:CluCut1}
\end{figure*}

\begin{figure}[h]
\includegraphics[width=0.45\textwidth]{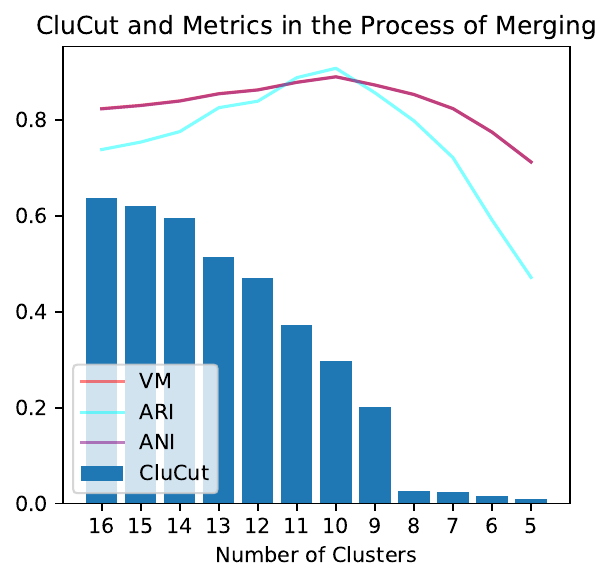}
\caption{Metrics in the Process of Merging (the USPS dataset)\\}
\label{Fig:CluCut2}
\end{figure}

\subsection{Experiments on CluCut-Guided Clustering Termination}\label{termination}

\noindent The $CluCut$ measure guide the termination of the cluster merging process in DPSM. To illustrate its functionality, we present a detailed account of processing the USPS dataset as we reduce the number of clusters from 16 to 5, as depicted in Fig. \ref{Fig:CluCut1}. 

When there are 16 clusters, each real class may be excessively divided into several clusters; however, no single cluster encompasses multiple real classes. Subsequently, the clusters are merged sequentially, ensuring that no merging occurs between clusters belonging to different classes. When reduced to 10 clusters, the clustering results closely align with the actual classification—this represents the outcome of DPSM given a specific number of clusters. After that, the approach starts to merge the two closest clusters.

The $CluCut$ measure, in conjunction with the three previously mentioned metrics, is illustrated as a graph in Fig. \ref{Fig:CluCut2}. 
We observe that $CluCut$ remains relatively high with no fewer than nine clusters, indicating that the method deems these mergers to be significant. Conversely, when the number of clusters falls below nine, $CluCut$ experiences a sharp decline, suggesting that the algorithm regards any further merges as unnecessary and advises against their execution. Fig. \ref{Fig:CluCut1} further illustrates this point. Prior to the formation of 9 clusters, there exist numerous connections among the merged clusters; conversely, after establishing 9 clusters, these merged clusters encompass the actual class.

The cluster metrics also reached a peak near 9 clusters as shown in Fig. \ref{Fig:CluCut2}. Notably, the two metrics, VM and ANI, exhibit nearly identical values in DPSM, whereas they differ significantly in other methods. This finding indicates that our approach effectively captures the underlying structure of the data and further emphasizes its stability and repeatability.

\section{Conclutions}
\noindent The density determination technique based on the propagation process that we propose relies solely on the local neighborhood of nodes. This transforms the static graph structure into dynamic density information, effectively addressing the limitation of traditional density-based clustering methods in their application to graph structures. 

By leveraging this density information, we achieve a rigorous partitioning and demonstrate its completeness. The demonstration guarantees the existence of clusters, ensure that there is only one class of nodes, stabilize the merge process, and keep the overall method at a low time complexity of $O(ndk \log n + nkt)$.

Inspired by spectral clustering, we propose a $CluCut$ measure that functions on clusters. This measure has undergone significant improvements to guide the merging of smaller clusters, leading to the development of DPSM capable of autonomously determining the number of clusters.

With a series of experiments, we validate the above conclusions by summarizing the following findings:

1) When the number of clusters in the dataset is unknown, DPSM can achieve better and more stable clustering results than other similar methods.

2) When the number of clusters in the dataset is known, DPSM not only improves the clustering result by this prior, but also improves the clustering result better than other clustering methods that require a given number of clusters.

3) The density derived from propagation effectively delineates the centers of clusters and the boundaries between them. This characteristic ensures that smaller clusters, which are segmented based on propagated density, achieve a high degree of homogeneity.

4) When the $CluCut$ measure experiences a sudden decline, we can conclude that a uniform and well-partitioned clustering result has been achieved. This typically signifies the completion of the clustering task.

%References
\bibliographystyle{IEEEtran}
\bibliography{citation}

\begin{thebibliography}{10}
\providecommand{\url}[1]{#1}
\csname url@samestyle\endcsname
\providecommand{\newblock}{\relax}
\providecommand{\bibinfo}[2]{#2}
\providecommand{\BIBentrySTDinterwordspacing}{\spaceskip=0pt\relax}
\providecommand{\BIBentryALTinterwordstretchfactor}{4}
\providecommand{\BIBentryALTinterwordspacing}{\spaceskip=\fontdimen2\font plus
\BIBentryALTinterwordstretchfactor\fontdimen3\font minus
  \fontdimen4\font\relax}
\providecommand{\BIBforeignlanguage}[2]{{%
\expandafter\ifx\csname l@#1\endcsname\relax
\typeout{** WARNING: IEEEtran.bst: No hyphenation pattern has been}%
\typeout{** loaded for the language `#1'. Using the pattern for}%
\typeout{** the default language instead.}%
\else
\language=\csname l@#1\endcsname
\fi
#2}}
\providecommand{\BIBdecl}{\relax}
\BIBdecl

\bibitem{pedrycz2005knowledge}
W.~Pedrycz, \emph{Knowledge-based clustering: from data to information
  granules}.\hskip 1em plus 0.5em minus 0.4em\relax John Wiley \& Sons, 2005.

\bibitem{kansal2018customer}
T.~Kansal, S.~Bahuguna, V.~Singh, and T.~Choudhury, ``Customer segmentation
  using k-means clustering,'' in \emph{2018 international conference on
  computational techniques, electronics and mechanical systems (CTEMS)}.\hskip
  1em plus 0.5em minus 0.4em\relax IEEE, 2018, pp. 135--139.

\bibitem{domingos2001mining}
P.~Domingos and M.~Richardson, ``Mining the network value of customers,'' in
  \emph{Proceedings of the seventh ACM SIGKDD international conference on
  Knowledge discovery and data mining}, 2001, pp. 57--66.

\bibitem{li2012using}
X.~Li and T.~Murata, ``Using multidimensional clustering based collaborative
  filtering approach improving recommendation diversity,'' in \emph{2012
  IEEE/WIC/ACM International Conferences on Web Intelligence and Intelligent
  Agent Technology}, vol.~3.\hskip 1em plus 0.5em minus 0.4em\relax IEEE, 2012,
  pp. 169--174.

\bibitem{shao2009music}
B.~Shao, D.~Wang, T.~Li, and M.~Ogihara, ``Music recommendation based on
  acoustic features and user access patterns,'' \emph{IEEE Transactions on
  Audio, Speech, and Language Processing}, vol.~17, no.~8, pp. 1602--1611,
  2009.

\bibitem{moshtaghi2011clustering}
M.~Moshtaghi, T.~C. Havens, J.~C. Bezdek, L.~Park, C.~Leckie, S.~Rajasegarar,
  J.~M. Keller, and M.~Palaniswami, ``Clustering ellipses for anomaly
  detection,'' \emph{Pattern Recognition}, vol.~44, no.~1, pp. 55--69, 2011.

\bibitem{probierz2022clustering}
B.~Probierz, J.~Kozak, and A.~Hrabia, ``Clustering of scientific articles using
  natural language processing,'' \emph{Procedia Computer Science}, vol. 207,
  pp. 3449--3458, 2022.

\bibitem{jain1999data}
A.~K. Jain, M.~N. Murty, and P.~J. Flynn, ``Data clustering: a review,''
  \emph{ACM computing surveys (CSUR)}, vol.~31, no.~3, pp. 264--323, 1999.

\bibitem{newman2004fast}
M.~E. Newman, ``Fast algorithm for detecting community structure in networks,''
  \emph{Physical Review E—Statistical, Nonlinear, and Soft Matter Physics},
  vol.~69, no.~6, p. 066133, 2004.

\bibitem{wang2011community}
F.~Wang, T.~Li, X.~Wang, S.~Zhu, and C.~Ding, ``Community discovery using
  nonnegative matrix factorization,'' \emph{Data Mining and Knowledge
  Discovery}, vol.~22, pp. 493--521, 2011.

\bibitem{lloyd1982least}
S.~Lloyd, ``Least squares quantization in pcm,'' \emph{IEEE transactions on
  information theory}, vol.~28, no.~2, pp. 129--137, 1982.

\bibitem{nie2021coordinate}
F.~Nie, J.~Xue, D.~Wu, R.~Wang, H.~Li, and X.~Li, ``Coordinate descent method
  for $ k $ k-means,'' \emph{IEEE Transactions on Pattern Analysis and Machine
  Intelligence}, vol.~44, no.~5, pp. 2371--2385, 2021.

\bibitem{wang2022discrete}
R.~Wang, J.~Lu, Y.~Lu, F.~Nie, and X.~Li, ``Discrete and parameter-free
  multiple kernel k-means,'' \emph{IEEE Transactions on Image Processing},
  vol.~31, pp. 2796--2808, 2022.

\bibitem{zhang1996birch}
T.~Zhang, R.~Ramakrishnan, and M.~Livny, ``Birch: an efficient data clustering
  method for very large databases,'' \emph{ACM sigmod record}, vol.~25, no.~2,
  pp. 103--114, 1996.

\bibitem{nie2020unsupervised}
F.~Nie, W.~Zhu, and X.~Li, ``Unsupervised large graph embedding based on
  balanced and hierarchical k-means,'' \emph{IEEE Transactions on Knowledge and
  Data Engineering}, vol.~34, no.~4, pp. 2008--2019, 2020.

\bibitem{szekely2005hierarchical}
G.~J. Szekely, M.~L. Rizzo \emph{et~al.}, ``Hierarchical clustering via joint
  between-within distances: Extending ward's minimum variance method,''
  \emph{Journal of classification}, vol.~22, no.~2, pp. 151--184, 2005.

\bibitem{ng2001spectral}
A.~Ng, M.~Jordan, and Y.~Weiss, ``On spectral clustering: Analysis and an
  algorithm,'' \emph{Advances in neural information processing systems},
  vol.~14, 2001.

\bibitem{shi2000normalized}
J.~Shi and J.~Malik, ``Normalized cuts and image segmentation,'' \emph{IEEE
  Transactions on pattern analysis and machine intelligence}, vol.~22, no.~8,
  pp. 888--905, 2000.

\bibitem{rodriguez2014clustering}
A.~Rodriguez and A.~Laio, ``Clustering by fast search and find of density
  peaks,'' \emph{science}, vol. 344, no. 6191, pp. 1492--1496, 2014.

\bibitem{xu2018dpcg}
X.~Xu, S.~Ding, M.~Du, and Y.~Xue, ``Dpcg: an efficient density peaks
  clustering algorithm based on grid,'' \emph{International Journal of Machine
  Learning and Cybernetics}, vol.~9, no.~5, pp. 743--754, 2018.

\bibitem{ren2020effective}
C.~Ren, L.~Sun, Y.~Yu, and Q.~Wu, ``Effective density peaks clustering
  algorithm based on the layered k-nearest neighbors and subcluster merging,''
  \emph{IEEE Access}, vol.~8, pp. 123\,449--123\,468, 2020.

\bibitem{lotfi2020density}
A.~Lotfi, P.~Moradi, and H.~Beigy, ``Density peaks clustering based on density
  backbone and fuzzy neighborhood,'' \emph{Pattern Recognition}, vol. 107, p.
  107449, 2020.

\bibitem{zhu2002learning}
X.~Zhu and Z.~Ghahramani, ``Learning from labeled and unlabeled data with label
  propagation,'' \emph{ProQuest number: information to all users}, 2002.

\bibitem{raghavan2007near}
U.~N. Raghavan, R.~Albert, and S.~Kumara, ``Near linear time algorithm to
  detect community structures in large-scale networks,'' \emph{Physical Review
  E—Statistical, Nonlinear, and Soft Matter Physics}, vol.~76, no.~3, p.
  036106, 2007.

\bibitem{page1999pagerank}
L.~Page, ``The pagerank citation ranking: Bringing order to the web,''
  Technical Report, Tech. Rep., 1999.

\bibitem{defferrard2016convolutional}
M.~Defferrard, X.~Bresson, and P.~Vandergheynst, ``Convolutional neural
  networks on graphs with fast localized spectral filtering,'' \emph{Advances
  in neural information processing systems}, vol.~29, 2016.

\bibitem{Hull1994ADF}
J.~J. Hull, ``A database for handwritten text recognition research,''
  \emph{IEEE Trans. Pattern Anal. Mach. Intell.}, vol.~16, pp. 550--554, 1994.

\bibitem{von2007tutorial}
U.~Von~Luxburg, ``A tutorial on spectral clustering,'' \emph{Statistics and
  computing}, vol.~17, pp. 395--416, 2007.

\bibitem{van2014accelerating}
L.~Van Der~Maaten, ``Accelerating t-sne using tree-based algorithms,''
  \emph{The journal of machine learning research}, vol.~15, no.~1, pp.
  3221--3245, 2014.

\bibitem{galil1991data}
Z.~Galil and G.~F. Italiano, ``Data structures and algorithms for disjoint set
  union problems,'' \emph{ACM Computing Surveys (CSUR)}, vol.~23, no.~3, pp.
  319--344, 1991.

\bibitem{van1976design}
P.~van Emde~Boas, R.~Kaas, and E.~Zijlstra, ``Design and implementation of an
  efficient priority queue,'' \emph{Mathematical systems theory}, vol.~10,
  no.~1, pp. 99--127, 1976.

\bibitem{synthetic_control_chart_time_series_139}
R.~Alcock, ``{Synthetic Control Chart Time Series},'' UCI Machine Learning
  Repository, 1999, {DOI}: https://doi.org/10.24432/C59G75.

\bibitem{dermatology_33}
N.~Ilter and H.~Guvenir, ``{Dermatology},'' UCI Machine Learning Repository,
  1998, {DOI}: https://doi.org/10.24432/C5FK5P.

\bibitem{zhang2012finding}
X.~Zhang, L.~Zhang, X.-J. Wang, and H.-Y. Shum, ``Finding celebrities in
  billions of web images,'' \emph{IEEE Transactions on Multimedia}, vol.~14,
  no.~4, pp. 995--1007, 2012.

\bibitem{nene1996columbia}
S.~A. Nene, S.~K. Nayar, H.~Murase \emph{et~al.}, ``Columbia object image
  library (coil-20),'' 1996.

\bibitem{Nene1996COIL100}
S.~Nene, S.~Nayar, and H.~Murase, ``{Columbia Object Image Library:
  COIL-100},'' Department of Computer Science, Columbia University, Tech. Rep.
  CUCS-006-96, February 1996.

\bibitem{van2008visualizing}
L.~Van~der Maaten and G.~Hinton, ``Visualizing data using t-sne.''
  \emph{Journal of machine learning research}, vol.~9, no.~11, 2008.

\bibitem{rosenberg2007v}
A.~Rosenberg and J.~Hirschberg, ``V-measure: A conditional entropy-based
  external cluster evaluation measure,'' in \emph{Proceedings of the 2007 joint
  conference on empirical methods in natural language processing and
  computational natural language learning (EMNLP-CoNLL)}, 2007, pp. 410--420.

\bibitem{hubert1985comparing}
L.~Hubert and P.~Arabie, ``Comparing partitions,'' \emph{Journal of
  classification}, vol.~2, pp. 193--218, 1985.

\bibitem{vinh2009information}
N.~X. Vinh, J.~Epps, and J.~Bailey, ``Information theoretic measures for
  clusterings comparison: is a correction for chance necessary?'' in
  \emph{Proceedings of the 26th annual international conference on machine
  learning}, 2009, pp. 1073--1080.

\bibitem{mcinnes2018umap}
L.~McInnes, J.~Healy, and J.~Melville, ``Umap: Uniform manifold approximation
  and projection for dimension reduction,'' \emph{arXiv preprint
  arXiv:1802.03426}, 2018.

\bibitem{fukunaga1975estimation}
K.~Fukunaga and L.~Hostetler, ``The estimation of the gradient of a density
  function, with applications in pattern recognition,'' \emph{IEEE Transactions
  on information theory}, vol.~21, no.~1, pp. 32--40, 1975.

\bibitem{ester1996density}
M.~Ester, H.-P. Kriegel, J.~Sander, and X.~Xu, ``Density-based spatial
  clustering of applications with noise,'' in \emph{Int. Conf. knowledge
  discovery and data mining}, vol. 240, no.~6, 1996.

\bibitem{bishop2006pattern}
C.~M. Bishop and N.~M. Nasrabadi, \emph{Pattern recognition and machine
  learning}.\hskip 1em plus 0.5em minus 0.4em\relax Springer, 2006, vol.~4,
  no.~4.

\bibitem{frey2007clustering}
B.~J. Frey and D.~Dueck, ``Clustering by passing messages between data
  points,'' \emph{science}, vol. 315, no. 5814, pp. 972--976, 2007.

\bibitem{arthur2006k}
D.~Arthur and S.~Vassilvitskii, ``k-means++: The advantages of careful
  seeding,'' Stanford, Tech. Rep., 2006.

\bibitem{murtagh2014ward}
F.~Murtagh and P.~Legendre, ``Ward’s hierarchical agglomerative clustering
  method: which algorithms implement ward’s criterion?'' \emph{Journal of
  classification}, vol.~31, pp. 274--295, 2014.

\end{thebibliography}

\begin{IEEEbiography}[{\includegraphics[width=1in,height=1.25in,clip,keepaspectratio]{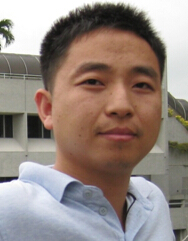}}]{Feiping Nie}
Feiping Nie received the Ph.D. degree in Computer Science from Tsinghua University, China in 2009, and is currently a full professor in Northwestern Polytechnical University, China. His research interests are machine learning and its applications, such as pattern recognition, data mining, computer vision, image processing and information retrieval. He has published more than 100 papers in the following journals and conferences: TPAMI, IJCV, TIP, TNNLS, TKDE, ICML, NIPS, KDD, IJCAI, AAAI, ICCV, CVPR, ACM MM. His papers have been cited more than 20000 times and the H-index is 99. He is now serving as Associate Editor or PC member for several prestigious journals and conferences in the related fields.
\end{IEEEbiography}

\begin{IEEEbiography}[{\includegraphics[width=1in,height=1.25in,clip,keepaspectratio]{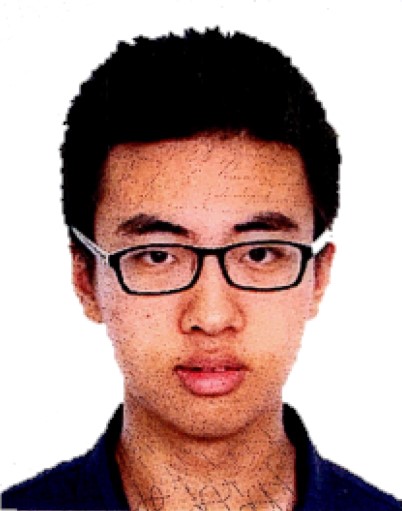}}]{Yitao Song}
Yitao Song received the B.S. from Northwestern Polytechnical University, Xi’an, China, in 2020. He is currently pursuing the Ph.D. degree with the School of Computer Science and the School of Artificial Intelligence, Optics and Electronics (iOPEN), Northwestern Polytechnical University, Xi’an, China.
His research interests include machine learning and data mining.
\end{IEEEbiography}

\begin{IEEEbiography}[{\includegraphics[width=1in,height=1.25in,clip,keepaspectratio]{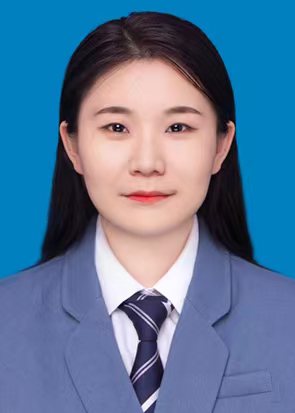}}]{Jingjing Xue}
Jingjing Xue received the Ph.D. degree from the School of Computer Science and the School of Artificial Intelligence, Optics and Electronics, Northwestern Polytechnical University, Xi’an, China, in 2023. She is currently a Lecturer with the School of Telecommunications Engineering, Xidian University, Xi’an. Her current research interests are machine learning and its applications, including matrix factorization, clustering, and model optimization. She is the author and coauthor of some scientific articles at top venues, including IEEE TPAMI, TNNLS, TCYB, TFS and TKDE.
\end{IEEEbiography}

\begin{IEEEbiography}[{\includegraphics[width=1in,height=1.25in,clip,keepaspectratio]{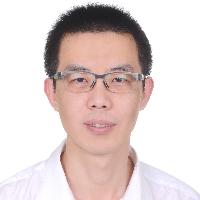}}]{Rong Wang}
Rong Wang is with the School of Artificial Intelligence, OPtics and ElectroNics (iOPEN), Northwestern Polytechnical University, Xi’an 710072, P.R. China, and also with the Key Laboratory of Intelligent Interaction and Applications (Northwestern Polytechnical University), Ministry of Industry and Information Technology, Xi’an 710072, P.R. China.
\end{IEEEbiography}

\begin{IEEEbiographynophoto}{Xuelong Li}
(Fellow, IEEE) is currently a Full Professor with the School of Artificial Intelligence, Optics and Electronics (iOPEN), Northwestern Polytechnical University, Xi’an, China.
\end{IEEEbiographynophoto}

\end{document}